\documentclass{article}
\usepackage[preprint]{acl}
\usepackage{times}
\usepackage{csquotes}

\usepackage{amsmath,amsfonts,bm}

\def\eqref#1{equation~\ref{#1}}

\def\1{\bm{1}}

\DeclareMathAlphabet{\mathsfit}{\encodingdefault}{\sfdefault}{m}{sl}
\SetMathAlphabet{\mathsfit}{bold}{\encodingdefault}{\sfdefault}{bx}{n}

\DeclareMathOperator*{\argmax}{arg\,max}

\usepackage{url}
\usepackage{soul}

\usepackage[utf8]{inputenc} 
\usepackage[T1]{fontenc}    
\usepackage{microtype,inconsolata}
\usepackage{times,latexsym}
\usepackage{graphicx} \graphicspath{{figures/}}
\usepackage{amsmath,amssymb,mathabx,mathtools,amsthm,nicefrac}
\usepackage[linesnumbered,ruled,vlined]{algorithm2e}
\usepackage{acronym}
\usepackage{enumitem}
\usepackage{balance}
\usepackage{xspace}
\usepackage{setspace}
\usepackage[skip=3pt,font=small]{subcaption}
\usepackage[skip=3pt,font=small]{caption}
\usepackage[capitalise,noabbrev,nameinlink]{cleveref}
\usepackage{booktabs,tabularx,colortbl,multirow,multicol,array,makecell,tabularray}
\usepackage{overpic,wrapfig}
\usepackage[misc]{ifsym}
\usepackage{pifont}
\usepackage{diagbox}

\makeatletter
\DeclareRobustCommand\onedot{\futurelet\@let@token\@onedot}
\def\@onedot{\ifx\@let@token.\else.\null\fi\xspace}

\makeatother

\makeatletter
\renewcommand{\paragraph}{%
  \@startsection{paragraph}{4}%
  {\z@}{0ex \@plus 0ex \@minus 0ex}{-1em}%
  {\hskip0em\normalfont\normalsize\bfseries}%
}
\makeatother

\crefname{algorithm}{Alg.}{Algs.}
\Crefname{algocf}{Algorithm}{Algorithms}
\crefname{section}{Sec.}{Secs.}
\Crefname{section}{Section}{Sections}
\crefname{table}{Tab.}{Tabs.}
\Crefname{table}{Table}{Tables}
\crefname{figure}{Fig.}{Figs.}
\Crefname{figure}{Figure}{Figures}
\crefname{equation}{Eq.}{Eqs.}
\Crefname{equation}{Equation}{Equations}
\crefname{appendix}{Appx.}{Appxs.}
\Crefname{appendix}{Appendix}{Appendices}

\definecolor{gblue}{HTML}{4285F4}
\definecolor{gred}{HTML}{DB4437}
\definecolor{ggreen}{HTML}{0F9D58}

\hypersetup{
}

\newcolumntype{P}[1]{>{\centering\arraybackslash}p{#1}}
\newcolumntype{M}[1]{>{\centering\arraybackslash}m{#1}}

\acrodef{metaicl-w}[Minnow]{Meta-training for IN-context learNing Of Words}
\acrodef{metaicl}[MetaICL]{Meta-training for In-Context Learning}
\acrodef{icl}[ICL]{in-context learning}
\author{%
  Wentao Wang$^1$ \hspace{1em} Guangyuan Jiang$^2$ \hspace{1em} Tal Linzen$^1$ \hspace{1em} Brenden M.\ Lake$^1$ \\
  $^1$New York University \hspace{1em} $^2$MIT \\
  \texttt{\{ww2135, linzen, brenden\}@nyu.edu} \hspace{1em} \texttt{jianggy@mit.edu}
}

\title{Rapid Word Learning Through Meta In-Context Learning}

\begin{document}

\maketitle

\begin{abstract}
Humans can quickly learn a new word from a few illustrative examples, and then systematically and flexibly use it in novel contexts.
Yet the abilities of current language models for few-shot word learning, and methods for improving these abilities, are underexplored.
In this study, we introduce a novel method, \ac{metaicl-w}.
This method trains language models to generate new examples of a word's usage given a few in-context examples, using a special placeholder token to represent the new word.
This training is repeated on many new words to develop a general word-learning ability.
We find that training models from scratch with \ac{metaicl-w} on human-scale child-directed language enables strong few-shot word learning, comparable to a large language model (LLM) pre-trained on orders of magnitude more data.
Furthermore, through discriminative and generative evaluations, we demonstrate that finetuning pre-trained LLMs with \ac{metaicl-w} improves their ability to discriminate between new words, identify syntactic categories of new words, and generate reasonable new usages and definitions for new words, based on one or a few in-context examples.
These findings highlight the data efficiency of \ac{metaicl-w} and its potential to improve language model performance in word learning tasks.
\end{abstract}

\section{Introduction}
\label{sec:intro}

Children can quickly learn a new word, or at least make meaningful inferences about its meaning, given only a few examples of its usage \citep{CareyBartlett1978,Bloom2000}.
For example, suppose a child who did not know the word \emph{ski} hears the following mentions of the word (without visual examples): ``\emph{Susie learned to ski last winter}'', ``\emph{People ski on tall mountains where there's lots of snow}'', and ``\emph{I saw Susie ski fast down the snowy mountain}.''
From these usage examples, the child might infer that \emph{ski} is a verb for a winter activity involving sliding down snowy mountains, and could begin understanding and using the word appropriately in new contexts.\footnote{Learning a new word is often equivalent to learning a new concept \citep{Murphy2002}. Therefore, we equate word learning to concept learning throughout the paper.}
This ability to generalize and use a new word in novel contexts from just a few examples reflects children's remarkable data efficiency in language learning, allowing them to quickly acquire vocabulary without requiring tens or hundreds of examples per word.


Compared to humans, current pre-trained language models are inefficient word learners, both in the total amount of pre-training data and the number of examples needed for each word.
Even though large language models (LLMs) are typically pre-trained on four or five orders of magnitude more language input than any single human could receive \citep{linzen2020accelerate,Frank2023BridgingTD}, they struggle with systematic generalizations of words that are rare or unseen in their training data \citep{wei-etal-2021-frequency,Razeghi2022ImpactOP,Kim2022UncontrolledLE,Batsuren2024EvaluatingST,Land2024FishingFM}.

This contrast between human learning and language model training raises two long-term research questions:
 1) Could language models develop a human-like ability for few-shot word learning without astronomical amounts of training data?
 2) Could existing LLMs be adapted to improve their few-shot word learning abilities, allowing them to systematically and flexibly use new words in new contexts?

\begin{figure*}[t]
\centering
\includegraphics[width=\textwidth]{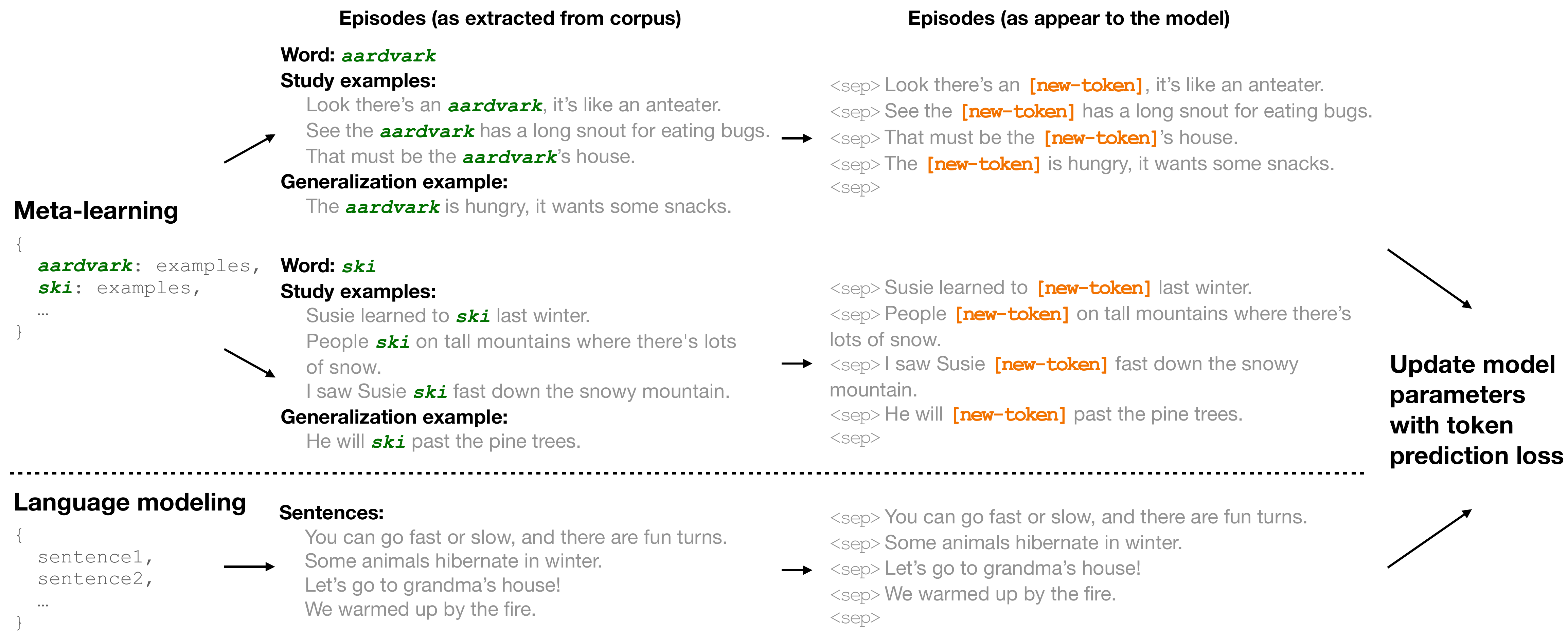}
\caption{Illustration of \ac{metaicl-w} (top) and language modeling (bottom), which can be mixed together during training such that both contribute to model updates. Each meta-learning episode in \ac{metaicl-w} aims to learn a \textcolor[RGB]{48,111,28}{new word} from a set of study examples (sentences that use the word) in the context and then generate a generalization example that also uses the word. Each language modeling episode contains a set of unrelated sentences without meta-learned words. An episode will be converted into a single sequence in which we replace the word to be learned (if it is a meta-learning episode) with a special placeholder token (e.g., \textcolor[RGB]{226,121,46}{\texttt{[new-token]}}) and concatenate/wrap the sentences with another special separator token (e.g., \texttt{<sep>}). We do gradient updates of the model parameters to optimize the next-token prediction loss on the sequence.}
\label{fig:method}
\end{figure*}

Here, we introduce a simple method, \acf{metaicl-w}, to train or finetune a language model to develop an in-context few-shot word learning capability (see Figure~\ref{fig:method} for an illustration of our method).
We adopt meta-training (i.e., meta-learning) since it has had successes in endowing neural networks with stronger systematic generalization, closely related to our objective of word learning (see \citealp{Russin2024Frege} for a review of the successes).
Specifically, we use \acl{metaicl} (\acs{metaicl}; \citealp{min-etal-2022-metaicl,chen-etal-2022-meta}) to train from scratch or finetune an auto-regressive language model to generate new usages of a new word given a set of illustrations of the new word in its previous context. \Acf{icl} builds and uses contextual representations of the new word on the fly without parameter updates. \acs{metaicl} repeats \ac{icl} on many different new words and optimizes the model parameters for a general word-learning ability.

To demonstrate the data efficiency of our method, we train language models from scratch with \ac{metaicl-w} using small datasets: a corpus of child-directed speech (CHILDES; \citealp{CHILDES}) and a corpus approximating the word count a child encounters during language acquisition (BabyLM-10M; \citealp{BabyLM}).
To foreshadow our results, we find that our method's performance on few-shot classification of new words from these datasets approaches that of the pre-trained \mbox{Llama-3 8B} \citep{Llama-3}, which was trained on vastly more data. This highlights how this ability can be developed from human-scale child-input data rather than the orders-of-magnitude larger datasets typically used to train LLMs.

We also finetune \mbox{Llama-3 8B} with \ac{metaicl-w} to see if we can enhance its word-learning ability.
In a series of discriminative and generative evaluations, we show that this improves \mbox{Llama-3 8B}'s ability to discriminate between new words, identify syntactic categories of new words, and generate reasonable new usages and definitions for new words, where each new word is learned from one or a few in-context examples.
Most of these improvements are achieved without specific training on these evaluation tasks.
We release our code at \url{https://github.com/wwt17/meta-learning-word}.

\section{Related Work}
\label{sec:related-work}

\subsection{The Rare Word Problem}
Word frequencies in natural corpora follow a highly skewed (Zipfian) distribution \citep{Zipf1949HumanBA}, resulting in a heavy tail of rare words. Additionally, new words are constantly entering the language \citep{Heaps1978InformationRC}.
To represent all possible words, various word-form-based methods have been proposed, including subword- and character-based tokenizations and using morphological information (see \citealp{Mielke2021BetweenWA} for a comprehensive survey).
However, representing a word alone does not help in learning it from a few contexts in which it occurs.
Models optimized for conventional language modeling still struggle with the usage of unfamiliar or completely novel words, tokens, or token sequences, where word-forms or token identities alone do not provide enough information \citep{Ott2018AnalyzingUI,Schick2020RareWA,wei-etal-2021-frequency,Razeghi2022ImpactOP,Kim2022UncontrolledLE,Batsuren2024EvaluatingST,Land2024FishingFM}.
Instead of representing new words based on word-forms, we discard word-form information and use a dedicated special placeholder token that is the same for every new word. In this way, we aim to develop a general and efficient ability to learn a word from a few contexts of its usage.

\subsection{Few-Shot Word Learning}
\label{sec:related-work-few-shot-word-learning}
Another line of previous work targets the problem of learning a new word from a few examples. 
Most previous work aims to produce a representation for the new word, i.e., an embedding, that fits into the global word embedding space so it can be used in the same way as other learned words \citep{Mikolov2013EfficientEO,Pennington2014GloVeGV}. The embedding can be produced by aggregating the embeddings of the contexts that the new word appears in \citep{Lazaridou2017MultimodalWM,khodak-etal-2018-la}, finetuning the embedding within the context \citep{herbelot-baroni-2017-high,Lampinen2017OneshotAF,hewitt2021initializing,kim-smolensky-2021-testing}, or utilizing the word-form information \citep{luong-etal-2013-better,schick2019learning}. 
More recent work uses Transformer layers to produce the embedding based on Word2Vec embeddings \citep[HiCE]{hu-etal-2019-shot}, or by aggregating similar embeddings of word contexts from a memory system \citep[Mem2Vec]{sun-etal-2018-memory}.
Also related to our approach, \citeauthor{Teehan2024CoLLEGeCE}'s \citeyearpar{Teehan2024CoLLEGeCE} work uses a meta-learning framework named CoLLEGe to train a Transformer encoder to produce an embedding for a new word from its examples of usage.
Our method also targets few-shot word learning, but is simpler than \citet{Teehan2024CoLLEGeCE} in architecture and training and does not produce a separate embedding for each new word.

\subsection{\acl{metaicl}}
Building on LLMs' \acl{icl} abilities \citep{GPT3}, \acf{metaicl} optimizes language models on multiple different tasks, each learned from a few in-context examples \citep{min-etal-2022-metaicl,chen-etal-2022-meta}.\footnote{MetaICL is different from \citet{cf-metaicl-2023}, which uses in-context learning instead of parameter updates to learn from multiple tasks.}
A class of tasks that \ac{metaicl} (or similar curriculums) aim to learn and generalize requires inferring the context-dependent mapping from the symbols to meanings \citep{Lake2023HumanlikeSG,huang2024lexinvariant,Anand2024DualPL,Park2024ICLR}.
We follow this work to use \ac{metaicl} for our word learning task, in which the mapping from a new word to its meaning should be inferred purely from its usage in the context.

\section{Method}
The goal of our method, \ac{metaicl-w}, is to enable a model to infer the meaning of a new word from a few examples of its usage so it can understand and generate novel usage examples of the word, coherently and systematically combining it with other words in new contexts.
To achieve this, \ac{metaicl-w} trains the model to generate another usage example of the new word---a task that, when sufficiently challenging, requires mastery of this ability.
\ac{metaicl-w} is a general framework that can be applied to both training a model from scratch and finetuning a pre-trained model.
After describing the method, we introduce the training data we use, a held-out word classification task for model evaluation and hyperparameter tuning, and how we use the off-the-shelf Llama model and the CoLLEGe model (introduced in Section~\ref{sec:related-work-few-shot-word-learning}) as baselines for our experiments.

\subsection{Method: \ac{metaicl-w}}
\label{sec:method}
Following the typical meta-learning approach, we construct episodes $\{\mathcal{T}_i\}_{i=1}^{N}$, each $\mathcal{T}_i$ consists of $K$ examples $\{x^{(i)}_k\}_{k=1}^{K}$ sampled in accordance with the desired task (Figure~\ref{fig:method}: top). In each episode, the model's task is to learn a new word $w_i$; each example $x^{(i)}_k$ is a sentence illustrating how $w_i$ is used. We concatenate the examples $\{x^{(i)}_k\}_{k=1}^{K}$ into a single sequence, separated by a special separator token (\texttt{<sep>} when training from scratch or a reserved special token in the \mbox{Llama-3 8B} vocabulary when finetuning \mbox{Llama-3 8B}). The objective is next-token prediction on this concatenated sequence: we expect the model to predict a new usage example given the previous examples, i.e., $p(x^{(i)}_k \mid x^{(i)}_1, \ldots, x^{(i)}_{k-1})$. We replace (mask) all occurrences of $w_i$ in the sequence with a special placeholder token (\texttt{[new-token]} when training from scratch or a different reserved special token when finetuning \mbox{Llama-3 8B}). The same placeholder token for the new word is shared across all episodes, such that the model does not learn a new embedding each time. Using the \emph{ski} example from Section~\ref{sec:intro}, the sequence for training models from scratch would be
\begin{displayquote}
\texttt{<sep>} Susie learned to \texttt{[new-token]} last winter \texttt{<sep>} People \texttt{[new-token]} on tall mountains where there's lots of snow \texttt{<sep>} I saw Susie \texttt{[new-token]} fast down the snowy mountain \texttt{<sep>}
\end{displayquote}
Note that our setting differs from previous \ac{metaicl} settings \citep{min-etal-2022-metaicl,chen-etal-2022-meta,Lake2023HumanlikeSG} in two ways. First, each example is not an input--output pair $(x^{(i)}_k, y^{(i)}_k)$, but just $x^{(i)}_k$. Second, there is no explicit separation between study examples and a query:\footnote{The study examples (or support examples) are the small number of examples given for a new episode from which the model learns or adapts. A query example (or generalization example) is a new example on which the model is tested after learning from the study examples. These terms are used in few-shot meta-learning literature (e.g., \citealp{Lake2023HumanlikeSG}).} our setting effectively uses every example $x^{(i)}_k$ as a query with all previous examples $x^{(i)}_1, \ldots, x^{(i)}_{k-1}$ as its study examples.

When we train a model from scratch, we also provide episodes of language modeling (without placeholder tokens) to further facilitate language learning, as illustrated in Figure~\ref{fig:method} (bottom). Each of these episodes consists of the same number of $K$ randomly sampled unrelated sentences, without new words. We concatenate them in the same format and train the model to perform next-token prediction on the concatenated sequences. Training batches of language modeling episodes interleave with the batches of meta-learning episodes. The model can determine whether an episode is for meta-learning or language modeling from whether the special placeholder token occurs in the first sentence.

\subsection{Data}
\label{sec:dataset}
\begin{figure*}[h]
\centering
\includegraphics[width=\textwidth]{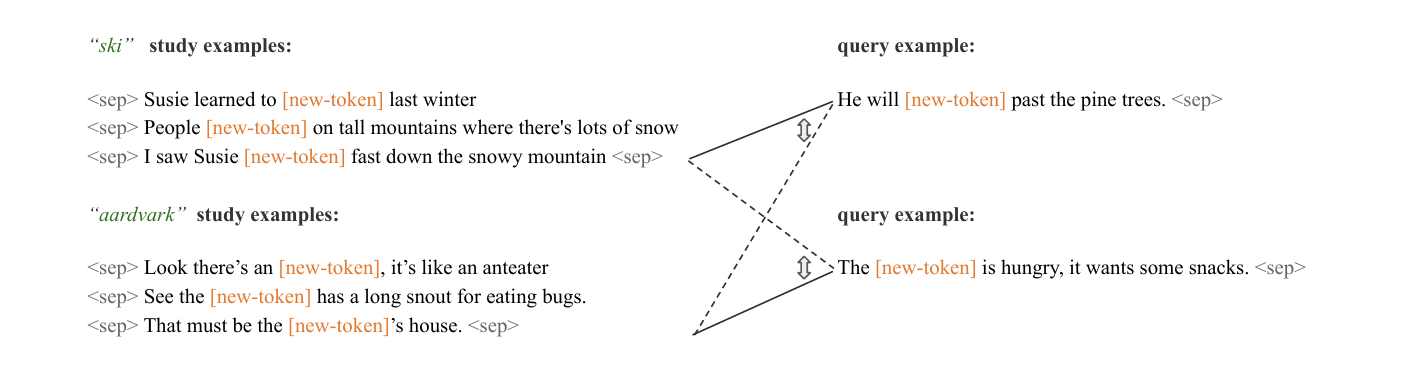}
\caption{An example task for held-out word classification. This is an example where we have $C = 2$ candidate words and $K = 4$ examples per word (left: three study examples per word; right: one query example per word). For each query example, we compute the conditional likelihood of its query sequence by prepending each context of study examples (lines in the middle), and we expect the correct context to give a higher likelihood (solid line) than the mismatched context (dashed line). See Appendix~\ref{app:word-classification} for a full description.}
\label{fig:word-classification-example}
\end{figure*}

To demonstrate the data efficiency of our method compared to humans, we use data sources that are close to children's language input in quantity or quality \citep{BabyLM}.
We construct one dataset from each of two corpora: CHILDES \citep{CHILDES} and BabyLM-10M \citep{BabyLM}.
CHILDES is a corpus of transcriptions of child--caregiver speech interactions. We use input to children (excluding utterances produced by children) in the North American English portion of CHILDES.
BabyLM is an English dataset including child-directed speech as well as additional data sources, such as children's books, transcriptions of dialogs between adults, and Wikipedia articles. We use the 10M word corpus constructed as part of the first BabyLM Challenge.

Each dataset consists of two disjoint components, one for meta-learning (the leftmost set in Figure~\ref{fig:method}: top) and the other for language modeling (the leftmost set in Figure~\ref{fig:method}: bottom).
We select a set of lower-frequency words in the corpus to be meta-learned in the meta-learning component.\footnote{Different word-forms of the same lexeme, like ``\emph{ski},'' ``\emph{skis},'' and ``\emph{skiing},'' are treated as different words in the dataset. See Appendix~\ref{app:word} for further discussion.}
Each meta-learned word $w$ has a set of $n_w$ sentence examples illustrating its usage.
We assign each sentence in the corpus to at most one meta-learned word, so the identity of the word masked by the placeholder token is not revealed in other meta-learning episodes.
During each training epoch, the $n_w$ examples for each word $w$ are split into $\lfloor\frac{n_w}{K}\rfloor$ (non-overlapping) episodes of $K$ examples, such that more frequent words have more episodes. This way of sampling episodes preserves the original Zipfian distribution of the word frequencies. Examples in the episodes are shuffled for each training epoch.
Other sentences in the corpus that have no meta-learned words are used for language modeling (Figure~\ref{fig:method} bottom).

We split both the meta-learning component (by word) and the language modeling component (by sentence) into training (80\%), validation (10\%) and test (10\%) portions.
Each dataset is used for both training models from scratch and finetuning pre-trained \mbox{Llama-3 8B}, but the text is formatted and tokenized differently (in addition to the different special tokens in Section~\ref{sec:method}; see Appendix~\ref{app:model} for the differences).
We provide additional details about data preprocessing, sentence assignment, dataset splitting, and text formatting in Appendix~\ref{app:dataset}, with statistics of our datasets shown in Table~\ref{tab:dataset-statistics}.
In the training portion, our CHILDES dataset contains 7,790 words to be meta-learned and has a total of $5.8$M tokens, while our BabyLM-10M dataset contains 15,821 words to be meta-learned and has a total of $7.8$M tokens.
In comparison, a child receives roughly $3$M to $12$M words per year \citep{Frank2023BridgingTD}, and thus our training data is of a similar magnitude to a year's worth of linguistic input for a child.

\subsection{Held-out Word Classification}
\label{sec:word-classification}

We introduce a word classification task, in which we measure the model's ability to discriminate the identities of new words that were never seen during training (i.e., held-out), based on in-context study examples. Validation accuracy on this task is used to tune training hyperparameters (e.g., learning rate; described later).

Given a query example sentence $q$ that uses a new word and a set of $C$ candidate words $\{w^{(c)}\}_{c=1}^{C}$, the task is to use the model likelihoods to match the query example to the most suitable one among the $C$ candidate words. Each $w^{(c)}$ is represented by a context containing a set of $K-1$ study examples $\{x^{(c)}_k\}_{k=1}^{K-1}$ illustrating its usage. (Note that in all query and study examples, the occurrences of the new word are replaced with the same special placeholder token, e.g., \texttt{[new-token]}, as described in Section~\ref{sec:method}.) The context of $w^{(c)}$ is a sequence in the same format as the first $K-1$ examples in a training episode, ending with a separator token (e.g., \texttt{<sep>}): \texttt{<sep>}~$x^{(c)}_1$~\texttt{<sep>}~$\cdots$~\texttt{<sep>}~$x^{(c)}_{K-1}$~\texttt{<sep>}. The query example is formatted as a continuation sequence of the context: $q$~\texttt{<sep>}.
This formatting ensures that concatenating a context sequence and a query sequence results in a sequence with $K$ examples, just like a sequence for a meta-learning training episode.
To determine the best match, we compute the conditional likelihood of the query sequence given the context: $p_\textrm{LM}(q \mid x^{(c)}_1, \ldots, x^{(c)}_{K-1})$. We then choose the word among the $C$ candidate words that gives the highest likelihood: $\argmax_{c} p_\textrm{LM}(q \mid x^{(c)}_1, \ldots, x^{(c)}_{K-1})$.
The choice is correct if it is the ground-truth word in the query $q$.

We evaluate each model (trained from scratch or finetuned) by measuring the classification accuracy on held-out meta-learned words from the validation or test portions of the model's training or finetuning corpus.
For each evaluation, we group $C$ distinct meta-learned words into a $C$-way classification task. For each word, we sample $K-1$ study examples and one query example to construct the task.
Figure~\ref{fig:word-classification-example} shows an example task.
See Appendix~\ref{app:word-classification} for additional details on task construction.

\subsection{Baselines}
\label{sec:baseline}

\subsubsection{Off-the-shelf Llama model}
\label{sec:baseline-llama}
For training models from scratch, we need an LLM that is pre-trained on massive data with conventional language modeling for data-efficiency comparison. To determine the effectiveness of finetuning an LLM, we need to evaluate its baseline word-learning ability.\footnote{As \citet{Anand2024DualPL} demonstrated, in vanilla language model training, in-context learning of new words is transient and eventually gives way to in-weights learning. Therefore, it is reasonable that the LLM's in-context word learning ability remains limited even after training on massive data.}
To address both needs, we use the off-the-shelf \mbox{Llama-3 8B} model as a baseline for word-learning tasks.
We experiment with both the pre-trained and the instruction-tuned variants of \mbox{Llama-3 8B}. We primarily report baseline results from the pre-trained variant, and present results from the instruction-tuned variant of \mbox{Llama-3 8B} only in the generative settings, where its performance may differ considerably from that of the pre-trained one.
For evaluation, we present a meta-learning episode to the Llama model in a text format similar to the training or finetuning sequences (Section~\ref{sec:method}), but designed to be more natural and closer to its pre-training data.
In particular, we use a pseudo-word (e.g., ``\textit{dax}'') as the placeholder for the new word, with a newline character and a star ``\verb|\n *|'' serving as the separator between examples, effectively formatting the examples as a list.\footnote{We choose the pseudo-word to be meaningless. However, a pre-trained LLM may ascribe a meaning to the pseudo-word based on its form. We acknowledge that replacing a word in an example with a pseudo-word could mislead the LLM and weaken the baseline. See Appendix~\ref{app:word} for detailed discussion.}
Using the \emph{ski} example in Section~\ref{sec:intro} again, the formatted text appears as follows:
\begin{displayquote}
 * Susie learned to \textit{dax} last winter

\nopagebreak
 * People \textit{dax} on tall mountains where there's lots of snow

\nopagebreak
 * I saw Susie \textit{dax} fast down the snowy mountain

\nopagebreak
 *
\end{displayquote}
The ``\verb|\n *|'' at the end serves as the last separator, like the last \texttt{<sep>} in the example sequence in Section~\ref{sec:method}.

\subsubsection{CoLLEGe}
\label{sec:baseline-college}
An alternative to \ac{metaicl-w} is to generate new embeddings for new words (Section~\ref{sec:related-work-few-shot-word-learning}).
For instance, CoLLEGe uses meta-learning to train an additional transformer encoder layer over a pre-trained MLM, RoBERTa \citep{RoBERTa}, to generate new input and output embeddings for a new word token (e.g., \texttt{[new-token]}) based on a set of study examples. The new token is then used by the pre-trained \mbox{Llama-2 7B} \citep{Llama-2}.
We use the original checkpoint of CoLLEGe as another baseline.
In the held-out word classification task (Section~\ref{sec:word-classification}), the conditional likelihood $p_\textrm{LM}(q \mid x^{(c)}_1, \ldots, x^{(c)}_{K-1})$ is computed by using only the input and output embeddings generated by CoLLEGe based on the study examples $x^{(c)}_1, \ldots, x^{(c)}_{K-1}$.
For fair comparison between \ac{metaicl-w} and CoLLEGe, we also finetuned from \mbox{Llama-2 7B} with \ac{metaicl-w} and compare this \ac{metaicl-w} model to the CoLLEGe model.
We also run off-the-shelf \mbox{Llama-2 7B} as an additional baseline, which is the same as described in Section~\ref{sec:baseline-llama}.

\section{Training Models From Scratch}
\label{sec:train-from-scratch}
In this section, we investigate whether models can develop the ability of few-shot word learning from human-scale input.
We use the GPT-NeoX transformer architecture \citep{GPT-NeoX} with configurations modified from Pythia-160M \citep{Pythia}.\footnote{We use an architecture with modern features such as relative positional encoding which may help in extrapolation to longer sequences and more examples. See Appendix~\ref{app:model} for details of our modifications.}
We use word-level tokenization. We exclude words with a frequency less than five from the vocabulary and replace them with \texttt{<unk>} tokens. We likewise remove the words that are to be meta-learned from this vocabulary and replace all of their occurrences in sentences other than their meta-learning episodes with \texttt{<unk>}. As mentioned in Section~\ref{sec:method}, the vocabulary also includes two special tokens: the placeholder token \texttt{[new-token]} and the separator token \texttt{<sep>}.

On each of the two datasets (CHILDES and BabyLM-10M) we train three models from scratch (i.e., the models are randomly initialized), each with $K=5$ examples per episode and a different random seed.
In each of the three runs, we choose the checkpoint with the lowest validation loss on the meta-learning objective.
Using one random seed, we fix the batch size and tune other training hyperparameters, including the learning rate and weight decay, for the best 4-way ($C=4$) held-out word classification accuracy on the validation portion of the dataset (the task was introduced in Section~\ref{sec:word-classification}).
We then apply the same training hyperparameters to the other seeds.
See Appendix~\ref{app:model} for detailed architecture configurations and training hyperparameters including batch size, learning rate (with scheduling), and weight decay.
In the following, we report mean accuracies of models across the three runs on the test portion of the dataset they were trained on.

\paragraph{Results}
Models trained from scratch on $K=5$ examples per episode sampled from CHILDES and BabyLM-10M achieve test accuracies of 72\% and 77\%, respectively, on the 4-way ($C=4$) classification task.
These results are substantially higher than random chance (25\%) and close to the 71\% and 78\% accuracies achieved by \mbox{Llama-3 8B} baseline (70\% and 78\% accuracies by \mbox{Llama-2 7B} baseline), which was pre-trained on orders of magnitude more data.
We provide results in additional settings, including experiments with $K=10$ examples on CHILDES and 8-way ($C=8$) classification, in Appendix~\ref{app:word-classification}, Table~\ref{tab:word-classification}. Across all settings, models trained from scratch consistently achieve accuracies well above chance and within a 3\% margin of the \mbox{Llama-3 8B} baseline.
These findings (on CHILDES in particular) demonstrate that few-shot word learning can be effectively acquired using our method, even with human-scale child-input data.

\section{Finetuning Pre-trained LLMs}
\label{sec:finetuning}

In this section, we test if our method can improve pre-trained LLMs' in-context few-shot word learning abilities.
We finetune \mbox{Llama-3 8B} (and \mbox{Llama-2 7B}) with \ac{metaicl-w} three times on the meta-learning component of BabyLM-10M, each run with $K=5$ examples per episode and a different random seed.\footnote{We focus on finetuning models on BabyLM-10M in this section, since it is more diversified and usually yields better results than CHILDES.}
We refer to the models finetuned with \ac{metaicl-w} as \ac{metaicl-w} models.
We do not include the language modeling components since the LLM already learned a large vocabulary and is capable of language modeling.
We finetune from both the pre-trained and instruction-tuned variants of \mbox{Llama-3 8B}, but we refer to the models finetuned from the pre-trained variant by default, same as for the off-the-shelf baseline (Section~\ref{sec:baseline-llama}).
We freeze all of the model's parameters except the input and output embeddings of these two special tokens.\footnote{See Appendix~\ref{app:other_capabilities} for its effect on other general capabilities.} We initialize the embeddings of these two special tokens as the mean of all other input/output embeddings \citep{hewitt2021initializing}.
We select the checkpoint for each run and tune the learning rate in the same way as when training from scratch, except that we do not apply weight decay (Section~\ref{sec:train-from-scratch}).
See Appendix~\ref{app:model} for more details on text formatting, tokenization, and training hyperparameters including batch size and learning rate (with scheduling).
In the following, we evaluate the \ac{metaicl-w} models and baselines on a series of tasks.

\subsection{Held-out Word Classification}
\label{sec:word-classification-finetuned}
We first evaluate models on the held-out word classification task (Section~\ref{sec:word-classification}).
Finetuning \mbox{Llama-3 8B} with \ac{metaicl-w} boosts the test 4-way ($C=4$) classification accuracy from the baseline level of 78\% to 87\% on BabyLM-10M (and from 71\% to 79\% on CHILDES).
We provide results for additional values of $K$ and $C$ and for models based on \mbox{Llama-2 7B} (including CoLLEGe) in Appendix~\ref{app:word-classification}, Table~\ref{tab:word-classification}; broadly, across all settings, the \ac{metaicl-w} model improves test accuracy by 8--10\% over the \mbox{Llama-3 8B} baseline, while the CoLLEGe model performs 3--15\% worse than the \mbox{Llama-2 7B} baseline.
These findings show that \ac{metaicl-w} finetuning effectively improves the pre-trained LLM's in-context few-shot word learning ability.

Despite these strong results, this task does not assess more fine-grained aspects of meaning that may not be apparent from discriminating an arbitrary set of words, and the semantic coherence of the usage contexts could be a shortcut utilized by the model (see Appendix~\ref{app:word-classification} for further discussion).
To address this, we provide the next analysis focusing on the syntactic categories of words.

\subsection{Syntactic Category Classification}
\label{sec:syntactic-category-classification}
In this evaluation, we test if models can differentiate words in different syntactic categories, a crucial feature for systematic generalization.
We follow the classification paradigm introduced in Section~\ref{sec:word-classification}.
We use the methodology of \citet{kim-smolensky-2021-testing} as well as the dataset they constructed from MNLI, a Natural Language Inference dataset \citep{MNLI}. The dataset focuses on four syntactic categories (noun, verb, adjective, and adverb) and tests the ability to differentiate each pair of categories. See Appendix~\ref{app:syntactic-category-classification} for details of the dataset.

In each instance of the classification task, we learn two new words $w^{(1)}$ and $w^{(2)}$ in different syntactic categories; the syntactic category of each new word $w^{(i)}$ is unambiguously signaled by a study example $x^{(i)}$ (replacing the word with the placeholder, e.g., \texttt{[new-token]}).
For example, say $w^{(1)}$ is a noun and $w^{(2)}$ is a verb:
\begin{enumerate}[label=(\arabic*)]
\item \textit{A \texttt{[new-token]} needs two people.} (for $w^{(1)}$)
\item \textit{She \texttt{[new-token]} at the group.} (for $w^{(2)}$)
\end{enumerate}
We test our models on query examples that use a word in one of the two categories, as in the following examples:
\begin{enumerate}[label=(\arabic*)]
\item \textit{Keep everyone else company by sitting in the \texttt{[new-token]}.} (expecting $w^{(1)}$)
\item \textit{The colonel \texttt{[new-token]} us to a hotel.} (expecting $w^{(2)}$)
\end{enumerate}
Note that, unlike the previous task, query examples are semantically unrelated to the study examples in this task, thus excluding the shortcut of semantic coherence.
Below, we report the mean accuracies across three runs.

\begin{table}[t]
\small
\begin{center}
\begin{tabular}{P{1.58cm}c|c}
\toprule
{\bf Variant} & {\bf Method} & {\bf Mean Acc.\ (\%)} \\
\midrule
\multirow{1}{=}{from scratch}
&  \ac{metaicl-w} & 77 \\
\midrule
\multirow{2}{=}{Llama-3 8B}
& baseline        & 66 \\
& +\ac{metaicl-w} & \textbf{83} \\
\midrule
\multirow{3}{=}{Llama-2 7B}
& baseline        & 69 \\
& +CoLLEGe        & \textbf{80} \\
& +\ac{metaicl-w} & \textbf{80} \\
\bottomrule
\end{tabular}
\end{center}
\caption{Mean accuracies of each model on the syntactic category classification task.
The random chance level accuracy is 50\%.
See Appendix~\ref{app:syntactic-category-classification} for fine-grained results.
In the top row, we show the result of the model trained from scratch with \ac{metaicl-w} on BabyLM-10M (Section~\ref{sec:train-from-scratch}).
In the other two ruled rows, we show models based on \mbox{Llama-3 8B} and \mbox{Llama-2 7B}, respectively.
We only have results for CoLLEGe based on \mbox{Llama-2 7B} because the original checkpoint is based on \mbox{Llama-2 7B}.
\ac{metaicl-w} accuracies are much higher than random chance and their corresponding baselines.}
\label{tab:syntactic-classification-summary}
\end{table}

\begin{table*}[t]
\small
\begin{center}
\begin{tabular}{p{8cm}p{4.3cm}p{0.8cm}}
\toprule
\textbf{Study Example Sentences} & \textbf{\ac{metaicl-w} Generated Examples} & \textbf{Word} \\
\midrule
$\bullet$ the first blacksmiths were \texttt{[new-token]}.
$\bullet$ many civilisations were in the area that is now turkey, like the \texttt{[new-token]}, the roman empire and the byzantine empire.
$\bullet$ spread of hepatoscopy and astrology to \texttt{[new-token]}, etruscans, greeks and romans and to china
$\bullet$ the first major empire in the area was the \texttt{[new-token]} (from the 18th century to the 13th century bce).
&
1. the \texttt{[new-token]} were a people who lived in the area of turkey.
2. perhaps the most famous and widely used alchemical symbol, first popularized by \texttt{[new-token]} alchemists, is the ouroboros.
& \textit{hittites}
\\
\bottomrule
\end{tabular}
\end{center}
\caption{New examples generated for a word from the BabyLM-10M test portion by the \ac{metaicl-w} model. The first one is generated by greedy decoding, and the second one by sampling with top-p=$0.92$.
The \ac{metaicl-w} model learns that \emph{hittites} is an ancient ethnic group. However, the greedy-decoded example copies the information (turkey) from the study example, while the sampled example makes seemingly plausible but factually incorrect generalizations (the earliest known ouroboros is found in ancient Egyptian text.)
}
\label{tab:babylm-generation-brief}
\end{table*}

\paragraph{Results}
Mean accuracies on this syntactic category classification task are summarized in Table~\ref{tab:syntactic-classification-summary}.
We first find that the \mbox{Llama-3 8B} baseline achieves 66\% accuracy on this task, which is higher than random chance (50\%), suggesting that it can infer the syntactic categories of new words in one shot and generalize them to novel contexts.
The \ac{metaicl-w} model improves \mbox{Llama-3 8B}'s accuracy to 83\%, an increase of 17\% points over the baseline.
Meanwhile, both \ac{metaicl-w} and CoLLEGe improve \mbox{Llama-2 7B}'s accuracy from 69\% to 80\%, which is an increase of 11\% points.
Fine-grained results from models finetuned with \ac{metaicl-w}, trained from scratch, and CoLLEGe are provided in Appendix~\ref{app:syntactic-category-classification}.
We find in all settings that the \ac{metaicl-w} model improves accuracy by 4--23\% compared to the baseline on all pairs of categories.
These results show that \ac{metaicl-w} finetuning effectively helps in learning the syntactic categories of new words and generalizing accordingly, and is comparable to CoLLEGe in improvements.
In addition, note that our models are not specifically finetuned on this syntactic category classification task and dataset, demonstrating the generality of the acquired word learning ability.

\subsection{New Usage Example Generation}
\label{sec:example-generation}

The two tests we have described so far evaluate models in a discriminative setting. Here, we quantitatively and qualitatively evaluate if models use the new word appropriately in a generative setting.
For a \ac{metaicl-w} model finetuned with $K$ examples per episode, we evaluate it by showing it $K-1$ in-context study examples, formatted as a sequence in the classification setting (Section~\ref{sec:word-classification}).
We ask the model to do what it was trained for: We prompt the model with this sequence of study examples, which ends with a separator token, so the model will continue the sequence by generating a new usage example, ending with another separator token as End-Of-Sequence.
For CoLLEGe, we generate a new example using the prompt ``A single example sentence using the word `\texttt{[new-token]}' (in one line):''.

We sample study examples from two datasets: the BabyLM-10M test portion in Section~\ref{sec:dataset} and the Chimera dataset \citep{Lazaridou2017MultimodalWM}.
The Chimera dataset was specifically constructed for few-shot word learning. It has 33 different new words for learning, each referring to a ``chimera'' concept, i.e., a mixture of two existing and related concepts (e.g., cello and bagpipe).
The usage examples of a new word are sentences using one of the components of the chimera, randomly extracted from a large corpus.
See Appendix~\ref{app:example-evaluation} for more details of the dataset.

For the quantitative evaluation, we compare a pair of new usage examples generated from \mbox{Llama-3 8B} baseline and a \ac{metaicl-w} model finetuned from it, or the CoLLEGe baseline and a \ac{metaicl-w} model finetuned from \mbox{Llama-2 7B}.
The comparison is simulated as a head-to-head competition following \citet{Teehan2024CoLLEGeCE}.
Specifically, we provide \mbox{GPT-4o} \citep{GPT-4o} the same $K-1$ study examples in a list format with a pseudo-word ``\textit{dax}'' as the placeholder for the word, as in the off-the-shelf baseline (without the last separator; Section~\ref{sec:baseline-llama}), followed by a question ``Which of the following is a better next example for the word `dax', or they tie?'' with three shuffled options, including the two generations and one ``Tie''. (See Appendix~\ref{app:comparing-generations} for prompting details.) The choice of \mbox{GPT-4o} decides whether and which one model wins the competition, or whether the models were tied in quality.
For the qualitative evaluation, we manually pick meta-learned words (Table~\ref{tab:babylm-generation-brief} and Appendix~\ref{app:example-evaluation}) and examine the syntactic correctness and semantic appropriateness of the generated examples.

\begin{table}[t]
\small
\begin{center}
\resizebox{\linewidth}{!}{
\begin{tabular}{P{1.49cm}c|P{1.13cm}P{1.13cm}|P{1.13cm}}
\toprule
 & & \multicolumn{2}{c|}{\bf New Usage Example} & {\bf Definition} \\
{\bf Variant} & {\bf Method} & {\bf BabyLM-10M test} & {\bf Chimera} & {\bf CoLLEGe-DefGen} \\
\midrule
\multirow{2}{=}{Llama-3 8B}
& baseline & 31 & 53 & 27 \\
& +\ac{metaicl-w} & \textbf{50} & 42 & \textbf{40} \\
\midrule
\multirow{2}{=}{Llama-3 8B Instruct}
& baseline & 40 & 44 & 32 \\
& +\ac{metaicl-w} & 46 & 48 & 37 \\
\midrule
\multirow{2}{=}{Llama-2 7B}
& +CoLLEGe        & 14 & 29 & 4 \\
& +\ac{metaicl-w} & \textbf{73} & \textbf{63} & \textbf{46} \\
\bottomrule
\end{tabular}
}
\end{center}
\caption{Percentages of wins of each model when comparing the generations from the pairs of models in each box, judged by \mbox{GPT-4o}.
In the top two ruled rows, we compare \mbox{Llama-3 8B} baseline (pre-trained to instruction-tuned) with a \ac{metaicl-w} model finetuned from that baseline (averaged across 3 runs).
In the bottom-most ruled row, we compare CoLLEGe with the \ac{metaicl-w} model finetuned from the pre-trained \mbox{Llama-2 7B} (averaged across 3 runs).
The left two datasets are for new usage example generation (Section~\ref{sec:example-generation}; each new usage example is generated by providing 4 study examples), and the right-most one is for definition generation (Section~\ref{sec:definition-generation}; each definition is generated by providing 3 study examples).
Each new example or definition is generated by greedy decoding.
We boldface significantly more preferred models ($p<.05$ in paired t-tests across 3 runs).
(Results of top-p sampled generations are shown in Table~\ref{tab:generative-quantitative-compare-top-p-gpt4o} in Appendix~\ref{app:comparing-generations}.)
The percentage of ties is the remaining after subtracting the win percentages of the two models.
On average, \mbox{GPT-4o} more frequently chooses the \ac{metaicl-w} model as the winner compared to the corresponding baseline model in all settings except for the pretrained \mbox{Llama-3 8B} on Chimera. The improvements by \ac{metaicl-w} on the instruction-tuned \mbox{Llama-3 8B} are not significant ($p>.1$).}
\label{tab:generative-quantitative-compare-greedy-gpt4o}
\end{table}

\paragraph{Results}
For the quantitative evaluation, Table~\ref{tab:generative-quantitative-compare-greedy-gpt4o} shows the percentages of wins of each of the baseline and the \ac{metaicl-w} model on both the BabyLM-10M test portion and Chimera. Across all settings, the \ac{metaicl-w} model wins more often on average than the corresponding baseline except for the pretrained \mbox{Llama-3 8B} on Chimera, demonstrating the improvement brought by \ac{metaicl-w} and its better performance compared to CoLLEGe.
For the qualitative evaluation, Table~\ref{tab:babylm-generation-brief} shows a word picked from the BabyLM-10M test portion along with its study and generated examples. See Appendix~\ref{app:example-evaluation} for additional examples from the BabyLM-10M test portion and Chimera and detailed analysis of the generations. A manual analysis of these generated examples reveals that the \ac{metaicl-w} model more often generates syntactically correct and semantically plausible new usage examples compared to the baseline, confirming that \ac{metaicl-w} finetuning improves the ability to understand and use a new word. Nevertheless, in several cases, the \ac{metaicl-w} model still shows obvious syntactic and factual errors and merely rewords the study examples.

\subsection{Definition Generation}
\label{sec:definition-generation}

\begin{table*}[t]
\small
\begin{center}
\begin{tabular}{p{2.7cm}l|cccc}
\toprule
\multicolumn{2}{c|}{\bf Model} & \multicolumn{2}{c}{\bf CoLLEGe-DefGen} & \multicolumn{2}{c}{\bf Oxford} \\
\bf Variant & \bf Method & \bf BERTScore F1 & \bf ROUGE-L & \bf BERTScore F1 & \bf ROUGE-L \\
\midrule
\multirow{2}{=}{Llama-3 8B}
           &             baseline          & 85.1 & 14.9 & 83.2 & 11.0 \\
           & +\ac{metaicl-w}               & 85.4 & 18.7 & \textbf{84.7} & \textbf{16.3} \\
\midrule
\multirow{2}{=}{Llama-3 8B Instruct}
                   &     baseline          & 85.3 & 17.6 & 83.6 & 12.5 \\
                   &+\ac{metaicl-w}        & \textbf{85.8} & \textbf{20.7} & \textbf{84.7} & \textbf{16.5} \\
\midrule\midrule
\multirow{2}{=}{Llama-2 7B}
                   &+CoLLEGe               & 84.0 & 16.3 & 83.3 & 14.1 \\
                   &+\ac{metaicl-w}        & 82.9 & \textbf{18.0} & 83.6 & \textbf{15.6} \\
\midrule\midrule
FLAN-T5 XL    &+DefInstr baseline          & 83.1 & 12.4 & 84.9 & 19.4 \\
\bottomrule
\end{tabular}
\end{center}
\caption{Quantitative evaluation of 1-shot generated definitions by comparing them with ground-truth definitions.
See Table~\ref{tab:definition-quantitative-1-shot-full} in Appendix~\ref{app:definition-evaluation} for results from all models.
We sample an example per word from CoLLEGe-DefGen.
All definitions are generated with greedy decoding.
``\mbox{FLAN-T5 XL} +DefInstr'' is a specialized definition-generation model from \citet{giulianelli-etal-2023-interpretable}.
``baseline'' means using a pseudo-word `\textit{wug}' as the placeholder.
Scores of \ac{metaicl-w} models (``+\ac{metaicl-w}'') are averaged across three runs.
Finetuning \mbox{Llama-3 8B} with \ac{metaicl-w} improves the baseline models on both datasets and both metrics ($p<.01$), and the \ac{metaicl-w} model finetuned from the instruction-tuned variant of \mbox{Llama-3 8B} performs the best on CoLLEGe-DefGen ($p<.01$; likely due to its better instruction-following ability; no significant difference is found on Oxford).
The \ac{metaicl-w} model beats CoLLEGe on ROUGE-L ($p=.027$ on CoLLEGe-DefGen and $p=.012$ on Oxford) but not on BERTScore F1 ($p=.289$ on Oxford).
The specialized model (``\mbox{FLAN-T5 XL} +DefInstr'') performs the best on Oxford ($p<.01$), but note that it is specialized in definition generation and was finetuned on Oxford.
}
\label{tab:definition-quantitative-1-shot}
\end{table*}

\begin{table*}[t]
\small
\begin{center}
\begin{tabular}{p{4.9cm}p{2.9cm}p{4.1cm}p{1.4cm}}
\toprule
\textbf{Example Sentence} & \textbf{\ac{metaicl-w} Definition} & \textbf{True Definition} & \textbf{Word}
\\
\midrule
Despite his greed, the businessman felt bound by a \texttt{[new-token]} to maintain ethical practices.
& a promise or agreement to do something
& a moral obligation or command that is unconditionally and universally binding
& \textit{categorical imperative}
\\
\bottomrule
\end{tabular}
\end{center}
\caption{Definition for a word from CoLLEGe-DefGen generated by the \ac{metaicl-w} model finetuned from instruction-tuned \mbox{Llama-3 8B} with greedy decoding.
The definition is generated using the single example sentence shown and provided in context.
The generated definition managed to infer the core semantic features from the examples, though they are not precise enough compared to the true definitions.
In the example, the \ac{metaicl-w} definition for ``\textit{categorical imperative}'' captures the core meaning of obligation, which is a reasonable contrast to the businessman's greed, but misses the ``unconditionally and universally binding'' aspect in the true definition.
}
\label{tab:defgen-definition}
\end{table*}

To further probe how well \ac{metaicl-w} finetuning helps the model understand a new word, we prompt each model to generate a definition for the word given one or a few usage examples.
We again follow \citet{Teehan2024CoLLEGeCE} for definition generation and evaluation, as well as the two evaluation datasets they used: CoLLEGe-DefGen, which they created, and the Oxford dataset \citep{gadetsky-etal-2018-conditional}.
CoLLEGe-DefGen was constructed by selecting 954 words from WordNet \citep{WordNet} and prompting \mbox{GPT-4} \citep{GPT4} to generate one definition and five usage examples for each word.
The model generates a definition from one, two, or three usage examples sampled for each word in this dataset (i.e., 1-, 2-, or 3-shot).
The Oxford test set consists of 12,232 words, each with a definition and a usage example collected from the Oxford Dictionary.
The model generates a definition from the only usage example for each word in this dataset (i.e., 1-shot).
To generate a definition, we prompt Llama and \ac{metaicl-w} models with the sequence of the usage example(s) (as in Section~\ref{sec:example-generation}) followed by ``The word \texttt{[new-token]} in the above sentence(s) is defined as "''\footnote{The prompt ends with a double quotation mark ("), so that the model will continue with a definition ending at another double quotation mark. This makes extracting definition easy.} (\texttt{[new-token]} is instead the placeholder token or pseudoword, as appropriate).
For CoLLEGe, we use the same prompt but without the in-context usage example(s).
See Appendix~\ref{app:definition-evaluation} for details of data preprocessing and additional specialized definition-generation models from comparison \citep{giulianelli-etal-2023-interpretable}.

For the quantitative evaluation, we perform two types of comparison.
The first type compares the model-generated and ground-truth definitions for each word by computing BERTScore F1 \citep{zhang2019bertscore} and \mbox{ROUGE-L} \citep{lin-2004-rouge}.
The second type compares a pair of definitions generated from \mbox{Llama-3 8B} baseline and a \ac{metaicl-w} model finetuned from it, or the CoLLEGe baseline and a \ac{metaicl-w} model finetuned from \mbox{Llama-2 7B}. Similarly to what we did in Section~\ref{sec:example-generation}, we ask \mbox{GPT-4o} a question (without usage examples): ``Which of the following is a better definition for the word `\textit{Word}', or they tie?'' where \textit{Word} is the ground-truth word form, followed by three shuffled options including the two generated definitions and one ``Tie'' (see Appendix~\ref{app:comparing-generations} for detailed prompting settings).\footnote{We only perform this comparison on the CoLLEGe-DefGen dataset due to the large scale of the Oxford dataset.}
For the qualitative evaluation, we manually inspect 1-shot generated definitions for words from each dataset (presented in Table~\ref{tab:defgen-definition} and Tables~\ref{tab:defgen-definition-more}~and~\ref{tab:oxford-definition} in Appendix~\ref{app:definition-evaluation}).

\paragraph{Results}
For the quantitative evaluation, we first present the 1-shot scores of comparing the model-generated and ground-truth definitions for \mbox{Llama-3 8B} and CoLLEGe baselines, the \ac{metaicl-w} models, and one specialized model in Table~\ref{tab:definition-quantitative-1-shot}. In Appendix~\ref{app:definition-evaluation}, we present 1-shot scores for all models (Table~\ref{tab:definition-quantitative-1-shot-full}) and averaged 1-, 2-, and 3-shot results on CoLLEGe-DefGen (Table~\ref{tab:definition-quantitative-defgen}). \ac{metaicl-w} finetuning improves the \mbox{Llama-3 8B} baseline by 0.3--1.5 on BERTScore F1 and 3.1--5.3 on \mbox{ROUGE-L}. On CoLLEGe-DefGen, the \ac{metaicl-w} model finetuned from the instruction-tuned \mbox{Llama-3 8B} outperforms all other non-specialized models across all settings. On Oxford, the \ac{metaicl-w} models finetuned from both variants of \mbox{Llama-3 8B} perform comparably well, but they are inferior to the largest specialized model by 2.9 on \mbox{ROUGE-L}. However, note that our \ac{metaicl-w} finetuning is neither tailored for generating definitions nor using these definition datasets.
In Table~\ref{tab:generative-quantitative-compare-greedy-gpt4o},  the \ac{metaicl-w} model is more often favored over each corresponding baseline.

For the qualitative evaluation, Table~\ref{tab:defgen-definition} shows \ac{metaicl-w}-model-generated and ground-truth definitions for a word from CoLLEGe-DefGen (see Tables~\ref{tab:defgen-definition-more}~and~\ref{tab:oxford-definition} in Appendix~\ref{app:definition-evaluation} for additional examples from CoLLEGe-DefGen and Oxford).
In our manual analysis, we find that definitions generated by the \ac{metaicl-w} model often capture most of the word meanings, form reasonable inferences from the contexts, and outperform the baselines. However, they are not always precise compared to the ground-truth definitions.

\section{Conclusion}
In this work, we present \ac{metaicl-w}, a new method to improve language models' capability to learn a new word from a few in-context usage examples.
\ac{metaicl-w} successfully induced this ability in models trained from scratch with human-scale linguistic data, as indicated by their performances in differentiating new words (Section~\ref{sec:train-from-scratch}).
\ac{metaicl-w} finetuning further improved the word learning performance of a pre-trained LLM (\mbox{Llama-3 8B}), as demonstrated in their improvements in differentiating new words (Section~\ref{sec:word-classification-finetuned}~and~\ref{sec:syntactic-category-classification}) as well as in generating new usage examples (Section~\ref{sec:example-generation}) and definitions (Section~\ref{sec:definition-generation}) for the learned new words.
In summary, this word-learning capability enables models to systematically and flexibly understand and use a new word in novel contexts, and can be immediately transferred to other words and tasks without additional training.

The efficacy of \ac{metaicl-w}, or meta-learning in general, suggests that human-level efficiency in linguistic generalizations may be acquired through practicing over many instances of learning tasks, without presuming strict, explicit inductive biases \citep{Russin2024Frege,Irie2024NNPractice}.
Whether models achieve the generalizations in this work through human-like mechanisms, such as systematicity and categorical abstraction, remains for future analysis.

\section{Limitations} 


\paragraph{Learning Settings}
In this work, we consider word learning only in the text modality, in which the language model learns the meaning from the distribution of words. However, many words have real-world references, which usually accompany human word learning.
We also use aggregated data from multiple sources, not from single-human/child input.
Thus, a multimodal, grounded setting of word learning using a single agent's input would be more realistic.

In addition, we only consider learning a single new word on the fly, and each word is represented by the same special token.
However, in real-world learning, both humans and models need to continually learn multiple words, usages, and even abstract rules \citep{sinha-etal-2023-language,Lampinen2024,mueller-etal-2024-context}. Future work could implement continual learning of multiple different words by meta-training to learn real words or pseudo-words, and by utilizing long-term memory. In that way, we could also allow the rapid word learning ability to co-exist with other general abilities of language models.

\paragraph{Novelty of New Words When Testing LLMs}
When testing LLMs (Section~\ref{sec:finetuning}), the words and example sentences we use may already exist in the pre-training data, potentially allowing LLMs to recall known word meanings rather than learn genuinely new ones\footnote{\citet{eisenschlos-etal-2023-winodict} suggested a similar solution called the reverse dictionary \citep{hill2016PhraseDict}, through which the model may identify the underlying concept from the word definition.} (note, however, the Chimera dataset introduces new concepts which are unusual and not lexicalized).
The performance of the baseline LLMs shows that, even with this potential worry, there is room for improvement, which the \ac{metaicl-w}-finetuned LLMs are able to achieve.

Models trained from scratch with \ac{metaicl-w} do not have this limitation. Their training data explicitly excludes held-out test words (Section~\ref{sec:train-from-scratch}). Therefore, their test performance reflects their genuine ability to learn novel words, and this ability can be developed by \ac{metaicl-w}.

\paragraph{Morphological Features}
In the paper, we focus on learning word-forms, each represented by a single special placeholder token, without considering morphological features. However, morphological features in a word may allow systematic recombination of meanings from related words, decrease the time needed to derive and recognize the word \citep{NagyMorphologicalFamilies}, and help children to learn the word \citep{Moore2024}.
In Appendix~\ref{app:word}, we discuss in detail our treatments of words and potential issues.
In summary, our \ac{metaicl-w} models are somehow robust to these variations.
Future work could take morphological features into consideration during training, and conduct targeted evaluations of morphological variations of the learned words.

\paragraph{Quantitative Evaluation of Generations}
In the generative evaluation settings (Section~\ref{sec:example-generation}~and~\ref{sec:definition-generation}), we used ROUGE-L, BERTScore F1, and LLM-as-a-Judge (GPT-4o in our case) for automatic quantitative evaluations. However, these evaluations are imperfect.
For example, different metrics are suitable for generations in different lengths, and LLM evaluators are known to have biases and inconsistencies \citep{doostmohammadi-etal-2024-reliable,Stureborg2024LargeLM}. Future work should conduct careful human evaluations for further validation to avoid these potential issues.

\section*{Acknowledgements}
We thank Najoung Kim for providing the syntactic category classification dataset.
We thank Ryan Teehan for providing the code and checkpoints for the CoLLEGe model.
We also thank Michael Hu, Will Merrill, Sophie Hao, Byung-Doh Oh, Shauli Ravfogel, and other members of the Computation and Psycholinguistics Lab for insightful and helpful discussions and comments.
This work is supported by the National Science Foundation under NSF Award 1922658 (for Wentao Wang) and IIS-2239862.
This work is also supported in part through the NYU IT High Performance Computing resources, services, and staff expertise.

{
    \small
    \bibliography{ref,library_clean}
}


\clearpage
\appendix
\section{Word Usage Dataset Creation}
\label{app:dataset}
As we mentioned in Section~\ref{sec:dataset}, we construct one dataset from each of two corpora: CHILDES \citep{CHILDES} and BabyLM-10M \citep{BabyLM}.
The CHILDES dataset is licensed for use under a CC BY-NC-SA 3.0 license.\footnote{\url{https://talkbank.org/share/rules.html}} Our scientific use is under the terms of the license.\footnote{\url{https://creativecommons.org/licenses/by-nc-sa/3.0/}}
We did not find the license of the BabyLM dataset, which aggregated multiple public datasets. Since there is plenty of published work using this public dataset, we believe our scientific use does not violate any terms or conditions.
In the following, we describe how we preprocess these two corpora and create a word usage dataset from each corpus.
\paragraph{Preprocessing}
Since the basic units of our focus are words (as opposed to word pieces in other tokenization schemes), we need to identify words in the text.
To achieve this, we apply the same word-level tokenization to all datasets (for consistency) and mark word boundaries by whitespace during preprocessing.
Models trained from scratch use this word-level tokenization.
When the text is used in finetuning \mbox{Llama}, which comes with its pre-trained subword tokenizer, we remove the unnatural spaces introduced by the word-level tokenization and tokenize the text again with the \mbox{Llama} tokenizer, so the text format becomes closer to its pre-training data (See the Finetuning paragraph in Appendix~\ref{app:model} for further details of this process).
For CHILDES data, we preprocess the data in the same way as \citet{yedetore-etal-2023-poor} did, which uses children's input in the North American English portion,\footnote{The version of CHILDES data we use is different from that of \citet{yedetore-etal-2023-poor}, and the current version on the official webpage \url{https://childes.talkbank.org/access/Eng-NA/} has also changed from our version.} but we do not split and unk the data at the preprocessing stage.
For BabyLM data, we use the data in the 10M track of the BabyLM Challenge 2023, which mixes 10 portions, each from a different data source (child- or adult-oriented, speech transcription or written text like Wikipedia). We exclude the QED portion for its poor quality (also mentioned in the 2nd BabyLM Challenge). We apply word-level tokenization on untokenized portions, and then split the text into sentences using heuristics.
We use spaCy for all word-level tokenization along with Part-Of-Speech tagging.
We lowercase all text before preprocessing to unify the capitalization of words in different places.
We deduplicate sentences and remove sentences having less than 1 word (not counting punctuation).

\paragraph{Assigning sentences and splitting}
To create a dataset from a corpus, we first get the token frequencies of all words. (Here, a word means a word-form. We discuss its implications in Appendix~\ref{app:word}.)
Then we select the set of words to be meta-learned.
We will only consider nouns, verbs, adjectives, and adverbs to be meta-learned (a word's syntactic category is based on the word's most frequent Part-Of-Speech tag).
We choose two thresholds for meta-learned words: the maximum frequency of a meta-learned word and the minimum number of examples per meta-learned word.
We use a greedy algorithm to assign each sentence in the corpus to the example set of at most one potential meta-learned word that occurs in the sentence, so each meta-learned word has at least the minimum number of examples. This ensures that the model cannot infer the identity of the word masked by the placeholder token from other sentences. These words and their example sets constitute the meta-learning component of the dataset.
We include the remaining sentences not assigned to any meta-learned word in the language-modeling component.
Finally, we split both the meta-learning component (by word) and the language-modeling component (by sentence) into training (80\%), validation (10\%), and test (10\%) portions.

When training models from scratch, we build the vocabulary from the words occurring with a minimum frequency in the training portion (same as the minimum number of examples per meta-learned word) while excluding all meta-learned words. This ensures that meta-learned words, like the lowest-frequency words, are out-of-vocabulary and will be replaced by \texttt{<unk>} tokens, so they will never be learned in-weights.

Statistics of our created datasets are shown in Table~\ref{tab:dataset-statistics}.
Read our code for full details.

\begin{table*}[t]
\small
\begin{center}
\begin{tabular}{cc|ccc|ccc}
\toprule
& & \multicolumn{3}{c|}{\bf CHILDES} & \multicolumn{3}{c}{\bf BabyLM-10M} \\
\midrule
\multicolumn{2}{c|}{max.\ freq.\ of meta-learned words}  & \multicolumn{3}{c|}{200} & \multicolumn{3}{c}{15} \\
\multicolumn{2}{c|}{min.\ \#uses of meta-learned words}  & \multicolumn{3}{c|}{5} & \multicolumn{3}{c}{5} \\
\multicolumn{2}{c|}{vocabulary size}  & \multicolumn{3}{c|}{2179} & \multicolumn{3}{c}{22,696} \\
\midrule
\multicolumn{2}{c|}{portion}                                & training& valid. & test   & training & valid. & test   \\
\midrule
\multirow{6}{3em}{meta-learning}     & \#meta-learned words &    7790 &    973 &    975 &   15,821 &   1977 &   1979 \\
                                     & total \#uses         & 201,957 & 26,449 & 26,234 &  108,466 & 13,552 & 13,563 \\
                                     & mean \#uses          &   25.93 &  27.18 &  26.91 &     6.86 &   6.85 &   6.85 \\
                                     & total \#tokens       &1,899,159& 245,509& 243,387&2,072,560 &260,701 &257,933 \\
                                     & mean sentence length &    9.40 &   9.28 &   9.28 &    19.11 &  19.24 &  19.02 \\
                                     & unk rate             &  3.32\% & 3.28\% & 3.28\% &   3.61\% & 3.78\% & 3.91\% \\
\midrule
\multirow{4}{3em}{language modeling} & \#sentences          & 508,630 & 63,578 & 63,580 &  521,911 & 65,238 & 65,240 \\
                                     & total \#tokens       &3,927,120& 492,280& 490,990&5,721,893 &715,553 &715,111 \\
                                     & mean sentence length &    7.72 &   7.74 &   7.72 &    10.96 &  10.97 &  10.96 \\
                                     & unk rate             &  1.00\% & 1.03\% & 1.00\% &   1.44\% & 1.49\% & 1.47\% \\
\midrule
\multicolumn{2}{c|}{total \#tokens}                         &5,826,279& 737,789& 734,377&7,794,453 &976,254 &973,044 \\
\bottomrule
\end{tabular}
\end{center}
\caption{Dataset statistics. All statistics are based on tokens, which mostly correspond to words except punctuations due to our word-level tokenization. ``unk rate'' is the percentage of out-of-vocabulary tokens, which are replaced by \texttt{<unk>}, in all tokens. Unk rate is slightly higher in the validation and test portions than the training portion because we build the vocabulary from the training portion. As shown by the mean sentence lengths, the meta-learning sentences are longer on average than the language modeling sentences, since meta-learned words are of lower frequency and thus are usually in more complex sentences. We manually tune the two thresholds of meta-learned words so we have enough number of meta-learned words while the unk rate is not too high.}
\label{tab:dataset-statistics}
\end{table*}

\clearpage\clearpage
\section{Model and Training Configurations}
\label{app:model}
\paragraph{Training from scratch}
We slightly modify the configuration of \mbox{Pythia-160M} \citep{Pythia}, which uses the Transformer architecture GPT-NeoX \citep{GPT-NeoX}. The configuration has $12$ layers and a hidden dimension size of $768$.
We change the vocabulary size according to the corresponding dataset, as shown in Table~\ref{tab:dataset-statistics}.
We also include three special tokens in the vocabulary: the placeholder token \texttt{[new-token]}, the separator token \texttt{<sep>}, and \texttt{<unk>}, as mentioned in Section~\ref{sec:train-from-scratch}.
We change the Pythia configuration to tie the input and output embeddings. This makes the model parameter counts smaller, $86.7$M and $102.5$M for the model trained on CHILDES and BabyLM-10M, respectively.
For both models, we use batch size (i.e., number of episodes/sequences per batch) $8$ and AdamW optimizer \citep{Loshchilov2019} with initial learning rate $3 \times 10^{-4}$, and reduce the learning rate by multiplying $0.1$ when the validation loss has stopped improving for $2$ epochs. We apply weight decay $0.07$ and $0.15$ when training on the CHILDES and BabyLM-10M datasets, respectively. Other configurations, such as no dropout, are kept the same as \mbox{Pythia-160M}.
For each setting, we run $3$ times with random seed $\{0, 1, 2\}$.
Each run is performed on a single V100 GPU for 30 epochs (9--18 hours).

\paragraph{Finetuning}
We finetune \mbox{Llama-3 8B} \citep{Llama-3} and \mbox{Llama-2 7B} \citep{Llama-2} with \ac{metaicl-w} on each of the CHILDES and BabyLM-10M datasets, but we refer to the models finetuned on BabyLM-10M by default, as we mentioned in Section~\ref{sec:finetuning}.
We finetune from both the pre-trained and instruction-tuned variants of \mbox{Llama-3 8B}, but we refer to the models finetuned from the pre-trained variant by default, presenting results of finetuning from the instruction-tuned variant only in the generative settings, where their performance may differ considerably due to their different capabilities to follow the prompt. We finetune only the pre-trained variant of \mbox{Llama-2 7B} since that is what the CoLLEGe checkpoint is based on.
We use two reserved special tokens in the \mbox{Llama-3} tokenizer vocabulary (or two tokens added to the \mbox{Llama-2 7B} vocabulary) as the placeholder token and the separator token.
To make the tokenization more natural to the model's pre-training data, we clean up tokenization spaces in the text (e.g., the space before ``,'', ``.'', or ``'s'') introduced by the word-level tokenization during preprocessing and make the placeholder token absorbs any preceding spaces of the word.
Finetuning is minimally parameter-efficient: We finetune only the input and output embeddings of the two special tokens, while freezing all other parameters. Before finetuning, the input/output embedding of either token is initialized to the mean of all input/output embeddings \citep{hewitt2021initializing}.
We finetune models on CHILDES with $5$ or $10$ examples per episode, and on BabyLM-10M with $5$ examples per episode.
Detailed hyperparameters we use to finetune \mbox{Llama-3 8B} and \mbox{Llama-2 7B} are summarized in Table~\ref{tab:finetune-hparams}.
Other settings are the same as when training from scratch except that we do not apply weight decay.
Each run is performed on a single A100 GPU for 15 epochs on CHILDES (33 hours) or 12 epochs on BabyLM-10M (48 hours).

\begin{table}[t]
\small
\centering
\resizebox{\linewidth}{!}{
\begin{tabular}{cc|ccccc}
\toprule
\textbf{dataset} & $K$ & \makecell{\bf batch \\ \bf size} & \makecell{\bf max.\ seq. \\ \bf length} & \multicolumn{2}{c}{\bf initial learning rate} \\
\cmidrule{5-6}
        &   &   &   & \textbf{Llama-3 8B} & \textbf{Llama-2 7B} \\
\midrule
\multirow{2}{*}{CHILDES} 
        & 5  & 32 & 80  & $3 \times 10^{-3}$ & $1 \times 10^{-3}$ \\
\cmidrule{2-6}
        & 10 & 8  & 160 & $3 \times 10^{-4}$ & $1 \times 10^{-3}$ \\
\midrule
BabyLM-10M & 5 & 16 & 160 & $1 \times 10^{-3}$ & $3 \times 10^{-3}$ \\
\bottomrule
\end{tabular}
}
\caption{Finetuning hyperparameters for different datasets and settings.
$K$ is the number of examples per episode.
``batch size'' is the number of episodes/sequences per batch.
``max.\ seq.\ length'' is the maximum number of tokens we truncate the sequence to in order to control the memory usage.}
\label{tab:finetune-hparams}
\end{table}

\clearpage\clearpage
\section{Held-out Word Classification}
\label{app:word-classification}

\paragraph{An example task}
Here we provide a full explanation of the example task in Figure~\ref{fig:word-classification-example} to further explain the classification paradigm in Section~\ref{sec:word-classification}. Assume $K=4,~C=2$, and we reuse the example words and sentences in Figure~\ref{fig:method}. As Figure~\ref{fig:word-classification-example} shows, the word ``\emph{ski}'' has its three study examples concatenated into a sequence:
\begin{displayquote}
\texttt{<sep>} Susie learned to \texttt{[new-token]} last winter \texttt{<sep>} People \texttt{[new-token]} on tall mountains where there's lots of snow \texttt{<sep>} I saw Susie \texttt{[new-token]} fast down the snowy mountain \texttt{<sep>}
\end{displayquote}
and a query example of the word ``\emph{ski}'' is formatted as:
\begin{displayquote}
He will \texttt{[new-token]} past the pine trees. \texttt{<sep>}
\end{displayquote}
The word ``\emph{aardvark}'' has its three study examples concatenated into another sequence:
\begin{displayquote}
\texttt{<sep>} Look there’s an \texttt{[new-token]}, it's like an anteater \texttt{<sep>} See the \texttt{[new-token]} has a long snout for eating bugs. \texttt{<sep>} That must be the \texttt{[new-token]}'s house. \texttt{<sep>}
\end{displayquote}
and a query example of the word ``\emph{aardvark}'' is formatted as:
\begin{displayquote}
The \texttt{[new-token]} is hungry, it wants some snacks. \texttt{<sep>}
\end{displayquote}
When classifying the word ``\emph{ski}'', we compare the conditional likelihood of its query example ``\textcolor[RGB]{48,111,28}{He will \texttt{[new-token]} past the pine trees. \texttt{<sep>}}'' in the following two sequences:
\begin{displayquote}
\texttt{<sep>} Susie learned to \texttt{[new-token]} last winter \texttt{<sep>} People \texttt{[new-token]} on tall mountains where there's lots of snow \texttt{<sep>} I saw Susie \texttt{[new-token]} fast down the snowy mountain \texttt{<sep>} \textcolor[RGB]{48,111,28}{He will \texttt{[new-token]} past the pine trees. \texttt{<sep>}}
\end{displayquote}
\begin{displayquote}
\texttt{<sep>} Look there’s an \texttt{[new-token]}, it's like an anteater \texttt{<sep>} See the \texttt{[new-token]} has a long snout for eating bugs. \texttt{<sep>} That must be the \texttt{[new-token]}'s house. \texttt{<sep>} \textcolor[RGB]{48,111,28}{He will \texttt{[new-token]} past the pine trees. \texttt{<sep>}}
\end{displayquote}
and we expect the conditional likelihood to be higher in the former sequence.

\paragraph{Task construction}
As we mentioned in Section~\ref{sec:word-classification}, we need different meta-learned words in the same group. Therefore, different from training, we sample only one episode of $K$ examples per word from the validation/test portions so we do not repeat the same word in a classification group. We also fix the shuffle order so all models are evaluated on the same classification task instances.
We experimented with training models with $K \in \{5, 10\}$ examples per episode on CHILDES and BabyLM-10M and evaluated each of them on the corresponding dataset with the same $K$ and $C \in \{4, 8\}$. Training models with $K = 10$ examples per episode on BabyLM-10M was unsuccessful because the concatenated sequence was too long, exceeding the GPU memory, so we do not have results in this setting.

\paragraph{Weaknesses of the task}
We are aware of the weaknesses of this task.
Discriminating a new word from an arbitrary set of other new words is a relatively weak test of word meaning learning.
The task could be easy simply because different words are used in very different contexts, so the conditional likelihood may reflect just the coherence of the usage contexts between study and query examples, not the meaning of the new word (we demonstrate this point by an additional baseline below where we present the model only the usage contexts without new words).
In addition, results from the task do not tell us what features of word meanings the model is learning.
Our syntactic category classification task addresses these concerns by focusing on the syntactic aspect and breaking the semantic coherence between study and query examples (Section~\ref{sec:syntactic-category-classification}).

Below, we describe two kinds of baselines we run on this task.

\paragraph{Baseline: pre-trained LLM learning a pseudo-word in context (\mbox{Llama-3 8B} or \mbox{Llama-2 7B} with `\textit{dax}')}
This is the baseline model introduced in Section~\ref{sec:baseline-llama}. We follow the format described there and additionally prepend a prompt to make the performance better: ``The following lines are lowercased example sentences using a new word `\textit{dax}' in random order, one per line:''. (We discuss the consequence of using a same pseudo-word in Appendix~\ref{app:word}.)
\paragraph{Additional Baseline: pre-trained LLM modeling the coherence of usage contexts (\mbox{Llama-3 8B} with `')}
This is the additional baseline to evaluate the effectiveness of utilizing just the coherence of the contexts, as we discussed above. We remove the new word from each example (equivalent to replacing the new word with an empty string), so only the usage context of each example is retained.

For these baselines, we also experimented with the instruction-tuned variant of \mbox{Llama-3 8B} but it performs worse on this task.

Table~\ref{tab:word-classification} shows all models' held-out word classification results on the test portions of CHILDES and BabyLM-10M datasets.

\begin{table*}[t]
\small
\begin{center}
\resizebox{\linewidth}{!}{
\begin{tabular}{ccc|ccccccc}
\toprule
\textbf{dataset} & $K$ & $C$ & \makecell{\bf \ac{metaicl-w} \\ \bf from scratch} & \makecell{\bf Llama-3 8B \\ \bf with `'} & \makecell{\bf Llama-3 8B \\ \bf with `\textit{dax}'} & \makecell{\bf Llama-3 8B \\ \bf +\ac{metaicl-w}} & \makecell{\bf Llama-2 7B \\ \bf with `\textit{dax}'} & \makecell{\bf Llama-2 7B \\ \bf +\ac{metaicl-w}} & \makecell{\bf Llama-2 7B \\ \bf +CoLLEGe} \\
\midrule
\multirow{4}{*}{CHILDES}
        & \multirow{2}{*}{$5$}
              &   4 & 72.3(1.6) & 58.33 & 71.09 & 79.1(0.5) & 70.06 & \textbf{79.4}(0.3) & 62.45 \\
        &     &   8 & 59.8(0.4) & 46.49 & 60.02 & 70.4(0.2) & 59.09 & \textbf{70.8}(0.3) & 47.93 \\
\cmidrule{2-10}
        & \multirow{2}{*}{$10$}
              &   4 & 75.1(0.7) & 66.56 & 76.53 & \textbf{84.9}(0.2) & 76.23 & 82.0(0.4) & 63.80 \\
        &     &   8 & 63.4(1.5) & 56.17 & 66.05 & \textbf{75.9}(0.6) & 65.74 & 73.1(0.4) & 50.62 \\
\midrule
\multirow{2}{*}{BabyLM-10M}
        & \multirow{2}{*}{$5$}
              &   4 & 77.4(0.5) & 70.45 & 78.39 & \textbf{86.5}(0.6) & 78.34 & 85.8(0.5) & 75.25 \\
        &     &   8 & 67.5(0.7) & 60.12 & 69.74 & \textbf{80.5}(1.0) & 69.53 & 79.5(0.4) & 63.56 \\
\bottomrule
\end{tabular}
}
\end{center}
\caption{Accuracy (\%) of held-out word classification on the CHILDES and BabyLM-10M test sets.
We show the mean and the standard deviation (in brackets) of 3 runs.
``\ac{metaicl-w} from scratch'' means models trained from scratch on the corresponding dataset.
``Llama-3 8B with `''' means the baseline model without prompt and remove the new word (i.e., replace the new word with an empty string).
``Llama-3 8B with `\textit{dax}''' or ``Llama-2 7B with `\textit{dax}''' means the baseline model with prompt learning the new word `\textit{dax}'.
We use $K-1$ study examples in this classification task, and models except the baselines are trained/finetuned on $K$ examples per training episode so they see the same number of examples during training and evaluation.
$C$ is the number of words in each group, so we will have $\lfloor \frac{n_\text{episodes}}{C} \rfloor$ groups.
Note that we discard the last batch of less than $C$ episodes, so the used numbers of episodes are slightly smaller.
Results of ``\mbox{Llama-3 8B} with `''' show that the coherence of the context already provides better-than-chance accuracy on this classification task.
Results of ``\mbox{Llama-3 8B} with `\textit{dax}''' show that the pre-trained LLM already performs well.
However, ``\mbox{Llama-3 8B} +\ac{metaicl-w}'' outperforms the baselines by a large margin, showing the effectiveness of our method.
Models finetuned with \ac{metaicl-w} from the instruction-tuned variant of \mbox{Llama-3 8B} perform worse than or close to the pre-trained variant here
(the instruction-tuned variant finetuned with \ac{metaicl-w} has 86.3\% (4-way) and 80.1\% (8-way) mean classification accuracies; the instruction-tuned variant with `\textit{dax}' has 75.2\% (4-way) and 66.0\% (8-way) classification accuracies),
so we do not include their results here.
Similar improvements by \ac{metaicl-w} is also shown on \mbox{Llama-2 7B} by comparing the baseline ``\mbox{Llama-2 8B} with `\textit{dax}''' and ``\mbox{Llama-2 8B} +\ac{metaicl-w}''.
Moreover, the CoLLEGe baseline (``\mbox{Llama-2 8B} +CoLLEGe'') performs even worse than the baseline.
}
\label{tab:word-classification}
\end{table*}

\clearpage\clearpage
\section{Syntactic Category Classification}
\label{app:syntactic-category-classification}
As we mentioned in Section~\ref{sec:syntactic-category-classification}, we use the methodology of \citet{kim-smolensky-2021-testing} and the dataset they constructed.
The dataset was constructed from MNLI, a Natural Language Inference dataset \citep{MNLI}.
The task is to discriminate between a pair of words in two different syntactic categories.
They consider 4 syntactic categories: noun, verb, adjective, and adverb.
Therefore, they have 6 pairs of categories for discrimination.
For each category pair, the dataset contains two signal contexts (one for each category; we use them as the study examples) and 200 test sentences using a word unambiguously in either category (100 for each category; we use them as the query examples).
The main difference between our approach and that of \citet{kim-smolensky-2021-testing} is that, instead of finetuning a new word embedding on each signal context, we apply in-context learning, using each signal context as an in-context study example of the new word.
Read \citet{kim-smolensky-2021-testing} for further details.

\begin{figure*}[t]
\centering
\includegraphics[width=0.95\textwidth]{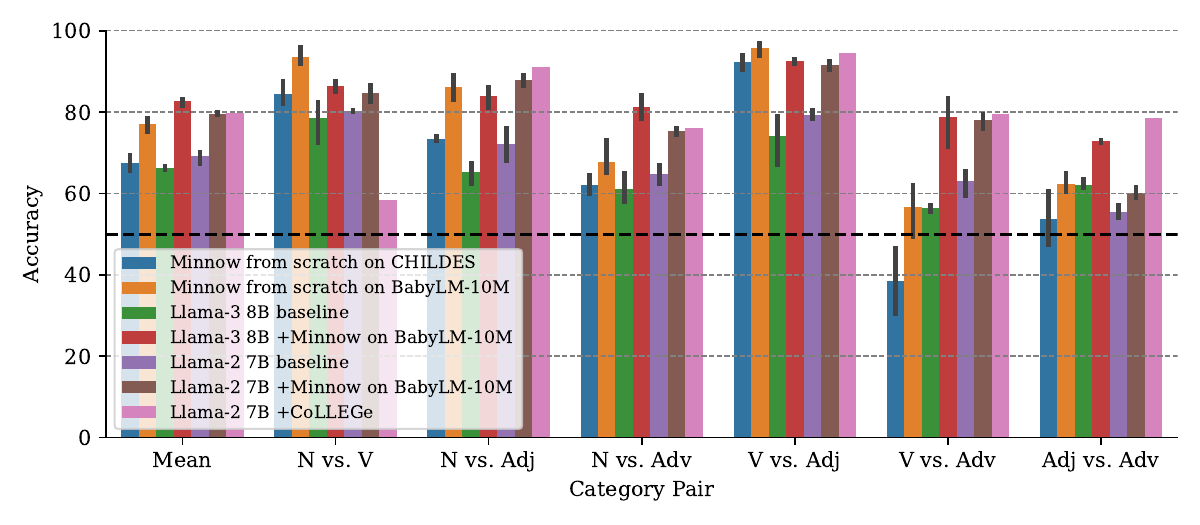}
\caption{Syntactic classification accuracy.
Error bar shows the 95\% confidence interval given 3 runs.
``\ac{metaicl-w} from scratch on CHILDES'' and ``\ac{metaicl-w} from scratch on BabyLM-10M'' mean the models trained from scratch with \ac{metaicl-w} on CHILDES and BabyLM-10M, respectively.
(These models have a closed vocabulary, so many words in the dataset will be Out-Of-Vocabulary and be presented as \texttt{<unk>}, which could make the task easier.)
``baseline'' means baseline with pseudo-word ``\emph{dax}'', ``\emph{wug}'', or ``\emph{blicket}''.
``+\ac{metaicl-w} on BabyLM-10M'' means \ac{metaicl-w} finetuning on BabyLM-10M.
``+CoLLEGe'' means the CoLLEGe model (\citealp{Teehan2024CoLLEGeCE}; generated new embeddings are used by \mbox{Llama-2 7B}).
``N'', ``V'', ``Adj'', and ``Adv'' are short for noun, verb, adjective, and adverb, respectively.
``Mean'' is the mean across all category pairs. The black dashed line marks the chance level (50\%).
``\mbox{Llama-3 8B} +\ac{metaicl-w} on BabyLM-10M'' shows improvement over ``\mbox{Llama-3 8B} baseline'' in all category pairs, with mean accuracy risen from 66\% to 83\%.
Meanwhile, both ``\ac{metaicl-w}'' and ``CoLLEGe'' improve the accuracy of ``\mbox{Llama-2 7B} baseline'' from 69\% to 80\%.
Note that ``\ac{metaicl-w} from scratch on BabyLM-10M'' has a 77\% mean accuracy, much better than the baseline accuracy and even comparable to the \ac{metaicl-w} models finetuned from \mbox{Llama-3 8B} on many category pairs, again demonstrating its data efficiency.}
\label{fig:syntactic-classification}
\end{figure*}

Results from models trained from scratch, \mbox{Llama-3 8B} and \mbox{Llama-2 7B} baseline, models finetuned from \mbox{Llama-3 8B} and \mbox{Llama-2 7B}, and the CoLLEGe baseline on the 6 category pairs and their mean are visualized in Figure~\ref{fig:syntactic-classification}.
Table~\ref{tab:category-classification-llama} shows detailed results from \mbox{Llama-3 8B} baseline and \mbox{Llama-3 8B} finetuned with \ac{metaicl-w} on BabyLM-10M.
Table~\ref{tab:category-classification-scratch} shows detailed results from models trained from scratch on both datasets.
Table~\ref{tab:category-classification-llama-2} shows detailed results from \mbox{Llama-2 7B} finetuned with \ac{metaicl-w} on BabyLM-10M and the CoLLEGe baseline.

\begin{table*}[t]
\small
\begin{center}
\begin{tabular}{ll|ccc|ccc}
\toprule
\multicolumn{2}{c|}{} & \multicolumn{3}{c|}{\bf Llama-3 8B baseline} & \multicolumn{3}{c}{\bf Llama-3 8B +\ac{metaicl-w}} \\
\bf Cat.\ 1 & \bf Cat.\ 2 & \bf Acc. & \bf Acc.\ (1$>$2) & \bf Acc.\ (2$>$1) & \bf Acc. & \bf Acc.\ (1$>$2) & \bf Acc.\ (2$>$1) \\
\midrule
Noun       & Verb       & 78.7(4.4) &    70.7(19.2) &    86.7(12.0) & \textbf{86.3}(1.5) & \textbf{74.7}(1.7) &          \textbf{98.0}(1.6) \\
Noun       & Adjective  & 65.2(2.1) &\textbf{87.3}(5.0)&  43.0(1.6) & \textbf{84.0}(2.2) &          71.3(4.6) & \textbf{96.7}(0.5) \\
Noun       & Adverb     & 61.0(2.9) &        32.3(4.5) &\textbf{89.7}(4.0) & \textbf{81.3}(2.2) & \textbf{75.7}(1.7) &          87.0(2.9) \\
Verb       & Adjective  & 74.2(5.2) &    88.0(11.3) &    60.3(18.1) & \textbf{92.7}(0.5) & \textbf{90.0}(2.2) & \textbf{95.3}(1.2) \\
Verb       & Adverb     & 56.3(0.6) &    53.7(12.3) &    59.0(11.9) & \textbf{78.8}(5.2) & \textbf{90.0}(2.4) & \textbf{67.7}(12.5) \\
Adjective  & Adverb     & 62.2(0.9) &    \textbf{60.3}(10.4) &    64.0(11.3) & \textbf{72.8}(0.2) & 57.3(2.6) & \textbf{88.3}(3.1) \\
\bottomrule
\end{tabular}
\end{center}
\caption{Comparing \mbox{Llama-3 8B} baseline and the \ac{metaicl-w} model finetuned from it on their accuracies (\%) of distinguishing two syntactic categories in novel contexts.
We show the mean and the standard deviation (in brackets) of 3 runs.
Following Table~1 in \citet{kim-smolensky-2021-testing}, `Acc.\ (1$>$2)' denotes the accuracy on the set of query sentences where Category 1 should be preferred over Category 2 (e.g., for row 1, assigning a higher likelihood to the noun-expecting query sentence when the placeholder represents a noun compared to a verb; using the examples in Section~\ref{sec:syntactic-category-classification}, the query example (1) expecting $w^{(1)}$ is among this set of query sentences, and it should have a higher likelihood when \texttt{[new-token]} represents $w^{(1)}$ compared to $w^{(2)}$.), and vice versa. Column `Acc.' lists the aggregate accuracy.
``\mbox{Llama-3 8B} baseline'' generally has accuracies better than chance (50\%) except for distinguishing certain pairs of categories. Additionally, ``\mbox{Llama-3 8B} +\ac{metaicl-w}'' improves over \mbox{Llama-3 8B} baseline in differentiating most category pairs, showing the effectiveness of finetuning with \ac{metaicl-w}.}
\label{tab:category-classification-llama}
\end{table*}

\begin{table*}[t]
\small
\begin{center}
\begin{tabular}{ll|ccc|ccc}
\toprule
\multicolumn{2}{c|}{} & \multicolumn{3}{c|}{\bf \ac{metaicl-w} from scratch on CHILDES} & \multicolumn{3}{c}{\bf \ac{metaicl-w} from scratch on BabyLM-10M} \\
\bf Cat.\ 1 & \bf Cat.\ 2 & \bf Acc. & \bf Acc.\ (1$>$2) & \bf Acc.\ (2$>$1) & \bf Acc. & \bf Acc.\ (1$>$2) & \bf Acc.\ (2$>$1) \\
\midrule
Noun      & Verb      & 84.5(2.3) &     79.7(3.7) &     89.3(4.5) & 93.5(1.8) &     90.0(2.2) &     97.0(1.4) \\
Noun      & Adjective & 73.5(0.4) &     50.7(2.9) &     96.3(2.1) & 86.2(2.5) &     79.7(5.4) &     92.7(1.9) \\
Noun      & Adverb    & 62.2(1.8) &     90.3(4.1) &     34.0(6.4) & 67.8(3.7) &     86.3(3.1) &     49.3(5.8) \\
Verb      & Adjective & 92.3(1.4) &     90.0(2.8) &     94.7(1.2) & 95.7(1.2) &     93.0(2.4) &     98.3(0.5) \\
Verb      & Adverb    & 38.5(6.5) &     57.3(14.7)&     19.7(1.7) & 56.7(5.3) &     68.7(5.8) &     44.7(11.4)\\
Adjective & Adverb    & 53.8(5.3) &     44.0(5.4) &     63.7(10.1)& 62.3(1.9) &     59.0(6.5) &     65.7(4.1) \\
\bottomrule
\end{tabular}
\end{center}
\caption{Accuracies (\%) of distinguishing two syntactic categories in novel contexts for models trained from scratch with \ac{metaicl-w}.
We show the mean and the standard deviation (in brackets) of 3 runs.
The formatting is the same as in Table~\ref{tab:category-classification-llama}.
Both models perform better than chance on many category pairs, suggesting that models can develop some ability to one-shot learn the syntactic category of a word from human-scale data with \ac{metaicl-w}.
In general, models trained on BabyLM-10M perform better than models trained on CHILDES, probably because the BabyLM dataset is more diverse and contains formal written texts, closer to the MNLI dataset, from which this test dataset is built.}
\label{tab:category-classification-scratch}
\end{table*}

\begin{table*}[t]
\small
\begin{center}
\begin{tabular}{ll|ccc|ccc}
\toprule
\multicolumn{2}{c|}{} & \multicolumn{3}{c|}{\bf Llama-2 7B +\ac{metaicl-w}} & \multicolumn{3}{c}{\bf Llama-2 7B +CoLLEGe} \\
\bf Cat.\ 1 & \bf Cat.\ 2 & \bf Acc. & \bf Acc.\ (1$>$2) & \bf Acc.\ (2$>$1) & \bf Acc. & \bf Acc.\ (1$>$2) & \bf Acc.\ (2$>$1) \\
\midrule
Noun       & Verb       & \textbf{84.7}(1.6) & 70.7(4.2) & \textbf{98.7}(0.9) & 58.5 & \textbf{92} & 25 \\
Noun       & Adjective  & 87.8(1.0) & 80.3(2.5) & 95.3(0.5) & \textbf{91.0} & \textbf{84} & \textbf{98} \\
Noun       & Adverb     & 75.3(0.6) & \textbf{70.0}(3.6) & 80.7(2.6) & \textbf{76.0} & 58 & \textbf{94} \\
Verb       & Adjective  & 91.5(0.8) & 95.0(1.4) & 88.0(2.2) & \textbf{94.5} & \textbf{99} & \textbf{90} \\
Verb       & Adverb     & 78.0(1.6) & \textbf{91.7}(1.2) & 64.3(2.1) & \textbf{79.5} & 68 & \textbf{91} \\
Adjective  & Adverb     & 60.2(1.0) & 45.0(4.3) & 75.3(4.9) & \textbf{78.5} & \textbf{69} & \textbf{88} \\
\bottomrule
\end{tabular}
\end{center}
\caption{Comparing the \ac{metaicl-w} models finetuned from \mbox{Llama-2 7B} and CoLLEGe \citet{Teehan2024CoLLEGeCE} on their accuracies (\%) of distinguishing two syntactic categories in novel contexts.
The formatting is the same as in Table~\ref{tab:category-classification-llama}.
Overall, both models perform well on the task (they both have 80\% mean accuracy as mentioned in Figure~\ref{fig:syntactic-classification}).
The CoLLEGe model performs better in more settings, but it fails to distinguish verbs from nouns (row 1, last column).}
\label{tab:category-classification-llama-2}
\end{table*}

\clearpage
\section{Comparing Generations}
\label{app:comparing-generations}

As we mentioned in Sections~\ref{sec:example-generation}~and~\ref{sec:definition-generation}, for the quantitative evaluation, we compare a pair of generations (new usage examples or definitions) from \mbox{Llama-3 8B} baseline and a \ac{metaicl-w} model finetuned from it, or the CoLLEGe baseline and a \ac{metaicl-w} model finetuned from \mbox{Llama-2 7B}.
In addition to \mbox{GPT-4o} evaluation, we have also asked the first author to conduct a small-scale human evaluation to compare \mbox{Llama-3 8B} baseline and a \ac{metaicl-w} model finetuned from it.\footnote{We acknowledge this is very limited and have problems, and leave larger-scale systematic human evaluation for future work.} We show \mbox{GPT-4o} and the human the same prompts.

For new usage example generation (Section~\ref{sec:example-generation}), we show \mbox{GPT-4o} or human the following text format:
\begin{displayquote}
The following lines are shuffled lowercased example sentences using a new word `dax', one per line:

* EXAMPLE-1

* EXAMPLE-2

* EXAMPLE-3

* EXAMPLE-4

Please answer in a single uppercase letter: Which of the following is a better next example for the word `dax', or they tie?

A) OPTION-A

B) OPTION-B

C) OPTION-C
\end{displayquote}
where OPTION-A, OPTION-B, OPTION-C are shuffled generation-1, generation-2, and ``Tie''.

For definition generation (Section~\ref{sec:definition-generation}), we do not have the examples (and the prompt before them) and instead have the direct prompt before the options: ``Please answer in a single uppercase letter: Which of the following is a better definition for the word `\textit{Word}', or they tie?'' where \textit{Word} is the ground-truth word form.

We always get the first letter (A, B, or C) of the \mbox{GPT-4o} response as the choice.

\paragraph{GPT-4o evaluation}
Tables~\ref{tab:generative-quantitative-compare-greedy-gpt4o}~and~\ref{tab:generative-quantitative-compare-top-p-gpt4o} show the results of comparing pairs of new usage examples or definitions generated from \mbox{Llama-3 8B} baseline (pre-trained to instruction-tuned) to a \ac{metaicl-w} model finetuned from it, or the CoLLEGe baseline and a Minnow model finetuned from \mbox{Llama-2 7B}, by greedy decoding and top-p=$0.92$, respectively.

\begin{table}[t]
\small
\begin{center}
\resizebox{\linewidth}{!}{
\begin{tabular}{P{1.49cm}c|P{1.13cm}P{1.13cm}|P{1.13cm}}
\toprule
 & & \multicolumn{2}{c|}{\bf New Example} & {\bf Definition} \\
\bf Variant & \bf Method & \bf BabyLM-10M test & \bf Chimera & \bf CoLLEGe-DefGen \\
\midrule
\multirow{2}{=}{Llama-3 8B}
& baseline & 37 & 46 & 24 \\
& +\ac{metaicl-w} & \textbf{53} & 42 & \textbf{31} \\
\midrule
\multirow{2}{=}{Llama-3 8B Instruct}
& baseline & 43 & 46 & 33 \\
& +\ac{metaicl-w} & 46 & 38 & 29 \\
\midrule
\multirow{2}{=}{Llama-2 7B}
& +CoLLEGe        & 5 & 11 & 5 \\
& +\ac{metaicl-w} & \textbf{85} & \textbf{68} & \textbf{30} \\
\bottomrule
\end{tabular}
}
\end{center}
\caption{Percentages of wins of each model when comparing the generations from the pairs of models in each box, judged by \mbox{GPT-4o}.
In the top two ruled rows, we compare \mbox{Llama-3 8B} baseline (pre-trained to instruction-tuned) with a \ac{metaicl-w} model finetuned from that baseline (averaged across 3 runs).
In the bottom-most ruled row, we compare CoLLEGe with the \ac{metaicl-w} model finetuned from the pre-trained \mbox{Llama-2 7B} (averaged across 3 runs).
The left two datasets are for new usage example generation (Section~\ref{sec:example-generation}; each new usage example is generated by providing 4 study examples), and the right-most one is for definition generation (Section~\ref{sec:definition-generation}; each definition is generated by providing 3 study examples).
Each new example or definition is generated by top-p=$0.92$.
We boldface significantly more preferred models ($p<.05$ in paired t-tests across 3 runs).
The percentage of ties is the remaining after subtracting the win percentages of the two models.
\mbox{GPT-4o} prefers the \ac{metaicl-w} model compared to the pre-trained \mbox{Llama-3 8B} baseline on two datasets.
\mbox{GPT-4o} also strongly prefers the \ac{metaicl-w} model compared to the CoLLEGe baseline ($p = 0.01$ on Chimera and $p < .001$ on other two datasets).
The difference made by \ac{metaicl-w} on the instruction-tuned \mbox{Llama-3 8B} are not significant ($p>.1$).}
\label{tab:generative-quantitative-compare-top-p-gpt4o}
\end{table}

\paragraph{Human evaluation}
Tables~\ref{tab:generative-quantitative-compare-greedy-human}~and~\ref{tab:generative-quantitative-compare-top-p-human} show the results of comparing pairs of new usage examples or definitions generated from \mbox{Llama-3 8B} baseline (pre-trained) to a \ac{metaicl-w} model finetuned from it, by greedy decoding and top-p=$0.92$, respectively.

\begin{table}[t]
\small
\begin{center}
\resizebox{\linewidth}{!}{
\begin{tabular}{P{1.49cm}c|P{1.13cm}P{1.13cm}|P{1.13cm}}
\toprule
 & & \multicolumn{2}{c|}{\bf New Example} & {\bf Definition} \\
\bf Variant & \bf Method & \bf BabyLM-10M test & \bf Chimera & \bf CoLLEGe-DefGen \\
\midrule
\multirow{2}{=}{Llama-3 8B}
& baseline & 6 & 9 & 20 \\
& +\ac{metaicl-w} & \textbf{32} & \textbf{30} & \textbf{22} \\
\bottomrule
\end{tabular}
}
\end{center}
\caption{Percentages of wins of each model when comparing the generations from \mbox{Llama-3 8B} baseline (pre-trained) with a \ac{metaicl-w} model finetuned from that baseline (with random seed 0), judged by the human. Each new example is generated by greedy decoding.
The percentage of ties is the remaining after subtracting the win percentages of the two models.
Due to the high cost of human evaluation, the human evaluates 50 pairs sampled from each of the BabyLM-10M test portion and the CoLLEGe-DefGen dataset.
The human more frequently choose the \ac{metaicl-w} model as the winner compared to the baseline.}
\label{tab:generative-quantitative-compare-greedy-human}
\end{table}

\begin{table}[t]
\small
\begin{center}
\resizebox{\linewidth}{!}{
\begin{tabular}{P{1.49cm}c|P{1.13cm}P{1.13cm}|P{1.13cm}}
\toprule
 & & \multicolumn{2}{c|}{\bf New Example} & {\bf Definition} \\
\bf Variant & \bf Method & \bf BabyLM-10M test & \bf Chimera & \bf CoLLEGe-DefGen \\
\midrule
\multirow{2}{=}{Llama-3 8B}
& baseline & 18 & 24 & 22 \\
& +\ac{metaicl-w} & \textbf{38} & \textbf{27} & \textbf{28} \\
\bottomrule
\end{tabular}
}
\end{center}
\caption{Percentages of wins of each model when comparing the generations from \mbox{Llama-3 8B} baseline (pre-trained) with a \ac{metaicl-w} model finetuned from that baseline (with random seed 0), judged by the human. Each new example is generated by sampling with top-p=$0.92$.
The percentage of ties is the remaining after subtracting the win percentages of the two models.
Due to the high cost of human evaluation, the human evaluates 50 pairs sampled from each of the BabyLM-10M test portion and the CoLLEGe-DefGen dataset.
The human more frequently choose the \ac{metaicl-w} model as the winner compared to the baseline.
}
\label{tab:generative-quantitative-compare-top-p-human}
\end{table}

To examine how the human and GPT-4o agree and disagree, Tables~\ref{tab:comparison-babylm-gpt-4o-human},~\ref{tab:comparison-chimera-gpt-4o-human}~and~\ref{tab:comparison-defgen-gpt-4o-human} show the counts of generations with each pair of human-GPT-4o judgments on the BabyLM-10M test portion, the Chimera dataset, and the CoLLEGe-DefGen dataset, respectively. In general, GPT-4o still underestimates the improvement of the \ac{metaicl-w} model compared to the baseline, enhancing the conclusion regarding the effectiveness of our method.
\begin{table}[t]
\small
\begin{center}
\begin{tabular}{c|ccc}
\toprule
\backslashbox{human}{GPT-4o} & +\ac{metaicl-w} & baseline & tie \\
\midrule
+\ac{metaicl-w} & 26 &  7 &  2 \\
baseline        &  3 &  9 &  0 \\
tie             & 20 & 23 & 10 \\
\bottomrule
\end{tabular}
\end{center}
\caption{Comparison between judgments made by \mbox{GPT-4o} and the human on the BabyLM-10M test portion. The human evaluates 50 words, each has two pairs of generations: one generated by greedy decoding and one generated by top-p=$0.92$ sampling, resulting in 100 pairs in total.
By comparing the off-diagonal numbers, we know that when the human and \mbox{GPT-4o} disagree, \mbox{GPT-4o} tends to favor the baseline, which suggests that \mbox{GPT-4o} still underestimates the improvement brought by finetuning with \ac{metaicl-w} compared to the human.}
\label{tab:comparison-babylm-gpt-4o-human}
\end{table}

\begin{table}[t]
\small
\begin{center}
\begin{tabular}{c|ccc}
\toprule
\backslashbox{human}{GPT-4o} & +\ac{metaicl-w} & baseline & tie \\
\midrule
+\ac{metaicl-w} & 17 &  0 &  2 \\
baseline        &  1 &  8 &  2 \\
tie             & 14 & 19 &  3 \\
\bottomrule
\end{tabular}
\end{center}
\caption{Comparison between judgments made by \mbox{GPT-4o} and the human on the Chimera dataset. There are 33 chimera words, each has two pairs of generations: one generated by greedy decoding and one generated by top-p=$0.92$ sampling, resulting in 66 pairs in total.}
\label{tab:comparison-chimera-gpt-4o-human}
\end{table}

\begin{table}[t]
\small
\begin{center}
\begin{tabular}{c|ccc}
\toprule
\backslashbox{human}{GPT-4o} & +\ac{metaicl-w} & baseline & tie \\
\midrule
+\ac{metaicl-w} & 15 &  4 &  6 \\
baseline        &  3 & 15 &  3 \\
tie             &  8 & 14 & 32 \\
\bottomrule
\end{tabular}
\end{center}
\caption{Comparison between judgments made by \mbox{GPT-4o} and the human on the CoLLEGe-DefGen dataset. The human evaluates 50 words, each has two pairs of generations: one generated by greedy decoding and one generated by top-p=$0.92$ sampling, resulting in 100 pairs in total.}
\label{tab:comparison-defgen-gpt-4o-human}
\end{table}

\clearpage\clearpage
\section{Evaluation of Generated New Usage Examples}
\label{app:example-evaluation}

As we mentioned in Section~\ref{sec:example-generation}, we sample study examples from two datasets: the BabyLM-10M test portion and the Chimera dataset \citep{Lazaridou2017MultimodalWM}.
Statistics of the BabyLM-10M test portion are in Table~\ref{tab:dataset-statistics}, Appendix~\ref{app:dataset}.
The Chimera dataset contains 33 chimeras.
A chimera is a mixture of two existing and related concepts (e.g., cello and bagpipe).
The usage examples of a chimera are sentences using one of the components of the chimera, randomly extracted from a large corpus.
For each chimera, we merge all examples in its 10 ``passages'' (each with a different pseudo-word in place of the word) in the original dataset, so we only have 33 different words to learn. We do not use the pseudo-words in the original dataset.

Tables~\ref{tab:babylm-generation}~and~\ref{tab:chimera-generation} show samples of example sentences from the BabyLM-10M test portion and the Chimera dataset along with the new usage examples generated by both the \mbox{Llama-3 8B} baseline and the \ac{metaicl-w} model finetuned from that baseline. Table~\ref{tab:babylm-generation-error} contains error analysis of new examples generated from the BabyLM-10M test portion, extending Table~\ref{tab:babylm-generation-brief} in Section~\ref{sec:example-generation}. For comparing the \ac{metaicl-w} model finetuned from \mbox{Llama-2 7B} and the CoLLEGe model on \mbox{Llama-2 7B}, we provide examples generated from the Chimera dataset and the BabyLM-10M test portion in Table~\ref{tab:chimera-generation-llama-2}~and~\ref{tab:babylm-generation-error-llama-2}, respectively.

\begin{table*}[t]
\small
\begin{center}
\begin{tabular}{p{5.3cm}p{2.3cm}p{2.3cm}p{3cm}p{0.8cm}}
\toprule
\bf Study Example Sentences & \bf \ac{metaicl-w} & \bf Baseline & \bf Judgment and Reason & \bf Word
\\
\midrule
$\bullet$ tales similar to " bluebeard " include " the fitcher's bird " by the brothers \texttt{[new-token]} and the english tale " mr. fox ".
$\bullet$ these types of tales include royalty and fantastical magic, many of which are published by brothers \texttt{[new-token]} and hans christian anderson.
$\bullet$ previously on \texttt{[new-token]}...
$\bullet$ the house in the wood from the german of \texttt{[new-token]}.
& previously in \texttt{[new-token]}
& the story is actually entitled " the bridal of trier " as mentioned on brothers \texttt{[new-token]} site.
& Baseline wins. \ac{metaicl-w} finetuned output simply changes the preposition of an example, which is likely incorrect. The baseline output copies ``brothers \texttt{[new-token]}'', which is likely correct.
& grimm
\\
\midrule
$\bullet$ men were sent to search the \texttt{[new-token]}, and there, on the very night before the deed was to be done, guy fawkes was found waiting till the time should come to set fire to the powder.
$\bullet$ the actual \texttt{[new-token]}   they've got some of the old er   writings that the, the monks   and you're not allowed to take photographs.
$\bullet$ searched all the tunnels and \texttt{[new-token]}.
$\bullet$ he puts them in \texttt{[new-token]}, anderson shelters, confined spaces, things like that.
& because they can light it in a \texttt{[new-token]} and blow it with the minimum amount of people.
& when a young teacher of history at the university of the commonwealth in london, \texttt{[new-token]} began the revolutionary research into indian philosophy of patanjali that was later to prove so fruitful.
& \ac{metaicl-w} model wins. The word likely refers to a place or space. Only the \ac{metaicl-w} finetuned output uses the word as a place.
& cellars
\\
\midrule
$\bullet$ the sorcerer wotan is using the \texttt{[new-token]} of aten to blot out the sun.
$\bullet$ unless it involves some kind of enchanted \texttt{[new-token]},
$\bullet$ i'll give her the lucky \texttt{[new-token]} you gave m
$\bullet$ this can be a long ritual, or it can take the form of simply wearing an \texttt{[new-token]} or other token, " against bad luck ".
& first, she goes with colm and feels a little guilty for keeping his \texttt{[new-token]}.
& the narrative treatment of \texttt{[new-token]} in the media exemplifies this process.
& \ac{metaicl-w} model wins. The word refers to a ritual token. The \ac{metaicl-w} finetuned output suggests it is an object, which could agree with the inference, while the baseline output does not convey a lot of information.
& amulet
\\
\midrule
$\bullet$ they are blind, their eyes having become reduced to \texttt{[new-token]} lenses under the skin, and they have no external ears, just a pair of tiny holes hidden under thick hair.
$\bullet$ it is a \texttt{[new-token]} reflex, a remnant of the past when pre - human babies clung to their mothers ' hair as the mothers swung through the trees.
$\bullet$ however, \texttt{[new-token]} structures may have their original function replaced with another.
$\bullet$ strong evidence for common descent comes from \texttt{[new-token]} structures.p397 the useless wings of flightless beetles are sealed under fused wing covers.
& the \texttt{[new-token]} structures of the human body are the appendix, the coccyx, the earlobes, the tailbone, and the wisdom teeth.
& the \texttt{[new-token]} structures of the human body are the same as those of other mammals.
& Tie. The word should be an adjective meaning an inherited biological feature that lost its original function. Both generations fit this interpretation. \mbox{GPT-4o} decides the \ac{metaicl-w} model as the winner.
& vestigial
\\
\midrule
$\bullet$ upon the very margin of the ditch, not thirty feet from where they crouched, an iron caldron \texttt{[new-token]} and steamed above a glowing fire ;
$\bullet$ all round the hull, in the blackness, the rippling current \texttt{[new-token]} and chattered like a little mountain stream.
$\bullet$ there hung another kettle that \texttt{[new-token]} and boiled ; but there was no fire beneath it, either.
$\bullet$ and when he lifted it up and looked down, there stood a great copper cauldron which \texttt{[new-token]} up and boiled away down there ;
& the boy looked down into the pot and saw that it \texttt{[new-token]} and boiled away.
& he turned his back upon them, and went in at the door in the \texttt{[new-token]}, which was a pleasant door to look upon.
& \ac{metaicl-w} model wins. The word should be a past-tense verb meaning a physical movement of an object, likely caused by steam or current. The \ac{metaicl-w} finetuned output is consistent with this meaning, but the baseline incorrectly uses the word as a noun.
& bubbled
\\
\bottomrule
\end{tabular}
\end{center}
\caption{New examples generated on study examples from the BabyLM-10M test portion by the \ac{metaicl-w} model and the \mbox{Llama-3 8B} baseline it is finetuned from, with greedy decoding.
In addition to \mbox{GPT-4o}'s judgment, the first author manually compared each pair of generations and gave the judgment and reason (Judgment and Reason).
When the author and \mbox{GPT-4o} compare the generations, they cannot see the ground-truth word but have to infer the possible meaning of the word.
When \mbox{GPT-4o}'s judgment disagrees with the author's, it is mentioned at the end of Judgment and Reason.
}
\label{tab:babylm-generation}
\end{table*}

\begin{table*}[t]
\small
\begin{center}
\begin{tabular}{p{5.5cm}p{2cm}p{2cm}p{3cm}p{1.3cm}}
\toprule
\bf Study Example Sentences & \bf \ac{metaicl-w} & \bf Baseline & \bf Judgment and Reason & \bf Word \\
\midrule
$\bullet$ Animals such as capybara, jaguars, jacare (\texttt{[new-token]}s) and hyacinth macaws are particularly vulnerable.
$\bullet$ Nadirpur stared at it as though it were a  \texttt{[new-token]}, his face quite drained.
$\bullet$ The owner will be summonsed to court for illegal possession of a \texttt{[new-token]} without a permit.
$\bullet$ I went to the cavern to get some, and found a \texttt{[new-token]} in there.
& $\bullet$ The \texttt{[new-token]} was a little more than a metre long, and had a thick, black coat.
& $\bullet$ The \texttt{[new-token]} was a large, black, furry creature with a long, pointed snout.
& \ac{metaicl-w} model wins. An alligator (inferred from the first example) is not furry.
& \mbox{alligator} + \mbox{rattlesnake}
\\
\midrule
$\bullet$ He is principal \texttt{[new-token]} of the English Baroque soloists, with whom he has recorded for Erato and Deutsche Grammophon.
$\bullet$ Banners point to the military aspect of \texttt{[new-token]} music: a Jacobite flag and the first Union flag.
$\bullet$ " Alexander Baillie said that he was impressed by her playing and even tried out her \texttt{[new-token]}!
$\bullet$ Where did the \texttt{[new-token]} idea at the end of Rubina's Blue Sky Happiness come from?
& $\bullet$ The first movement is a \texttt{[new-token]}, the second a minuet, and the third a rondo.
& $\bullet$ The \texttt{[new-token]} is a small, furry, four-legged creature with a long tail.
& Tie. Both generations are not for an instrument.
& \mbox{cello} + \mbox{bagpipe}
\\
\midrule
$\bullet$ 4. Allow the stock to settle then skim off the fat using \texttt{[new-token]} or fat separator.
$\bullet$ Add the pasta, cook it for about 12 minutes until just soft, then drain it in a \texttt{[new-token]}.
$\bullet$ Imagine you've got a bucketful of praise in every room of the house with a big \texttt{[new-token]}.
$\bullet$ Plants can be grown in tyres, old metal pots such as buckets, watering cans or even a \texttt{[new-token]}!
& $\bullet$ 2. Put the chicken in a \texttt{[new-token]} and cover with cold water.
& $\bullet$ The \texttt{[new-token]} is a very useful tool for the gardener.
& \ac{metaicl-w} model wins. The word refers to a kitchen utensil, likely a container. The baseline may be misled by the last example but do not notice the word ``even'' indicating the surprise of the usage.
& \mbox{ladle} + \mbox{colander}
\\
\midrule
$\bullet$ At about the same age I was in a cafe and I spilt some pop on my \texttt{[new-token]}s.
$\bullet$ It's also the first International home \texttt{[new-token]} in history that is reversible as well.
$\bullet$ Such as how if you have sufficiently-stretchy pants you could take them off without first removing your \texttt{[new-token]}s.
$\bullet$ It was in a cinema on 86th Street where he sat in the balcony, wearing an old sports jacket and open-necked \texttt{[new-token]}.
& $\bullet$ The \texttt{[new-token]} is a bit tight.
& $\bullet$ I was in a cafe and I spilt some pop on my \texttt{[new-token]}s.
& Tie. A word for clothes can be used in both example. \mbox{GPT-4o} decides the baseline as the winner.
& \mbox{trouser} + \mbox{shirt} \\
\bottomrule
\end{tabular}
\end{center}
\caption{New examples generated on study examples from the Chimera dataset \citep{Lazaridou2017MultimodalWM} by the \ac{metaicl-w} model and the \mbox{Llama-3 8B} baseline it is finetuned from, with greedy decoding.
Like in Table~\ref{tab:babylm-generation}, the first author provides the judgment and reason (Judgment and Reason).
The first example is a reasonable guess of the appearance of an animal.
The second example is related to music, but incorrectly infers the word to be a dance or musical form.
The last example correctly infers that the word is a kitchen utensil.
}
\label{tab:chimera-generation}
\end{table*}

\begin{table*}[t]
\small
\begin{center}
\begin{tabular}{p{5.5cm}p{2cm}p{2cm}p{3cm}p{1.3cm}}
\toprule
\bf Study Example Sentences & \bf \ac{metaicl-w} & \bf CoLLEGe & \bf Judgment and Reason & \bf Word \\
\midrule
$\bullet$ Animals such as capybara, jaguars, jacare (\texttt{[new-token]}s) and hyacinth macaws are particularly vulnerable.
$\bullet$ Nadirpur stared at it as though it were a  \texttt{[new-token]}, his face quite drained.
$\bullet$ The owner will be summonsed to court for illegal possession of a \texttt{[new-token]} without a permit.
$\bullet$ I went to the cavern to get some, and found a \texttt{[new-token]} in there.
& $\bullet$ The \texttt{[new-token]} was a big one, and it was a good thing i had a big knife.
& $\bullet$ 1. The \texttt{[new-token]} is a large, powerful, and dangerous animal.
& CoLLEGe wins. The \ac{metaicl-w} model's generation contains unrelated information.
& \mbox{alligator} + \mbox{rattlesnake}
\\
\midrule
$\bullet$ Fourteen-year-old Craig Ralphs and eighteen year old Glenn Cameron were subsequently arrested and charged with reckless driving and \texttt{[new-token]}s without consent.
$\bullet$ " I came in the same \texttt{[new-token]} as someone called Gleeson,"said Marcus.
$\bullet$ The chairman of \texttt{[new-token]}s has admitted the company is no longer a mass manufacturer.
$\bullet$ She had to continue driving the \texttt{[new-token]} all day, and again the following morning.
& $\bullet$ The \texttt{[new-token]} was a 1972 ford escort.
& $\bullet$ 1. The \texttt{[new-token]} is a large, heavy, and powerful bird.
& \ac{metaicl-w} wins. The \ac{metaicl-w} model's generation is very reasonable, while the CoLLEGe model's generation is incorrect.
& \mbox{car} + \mbox{van}
\\
\midrule
$\bullet$ He is principal \texttt{[new-token]} of the English Baroque soloists, with whom he has recorded for Erato and Deutsche Grammophon.
$\bullet$ Banners point to the military aspect of \texttt{[new-token]} music: a Jacobite flag and the first Union flag.
$\bullet$ " Alexander Baillie said that he was impressed by her playing and even tried out her \texttt{[new-token]}!
$\bullet$ Where did the \texttt{[new-token]} idea at the end of Rubina's Blue Sky Happiness come from?
& $\bullet$ The \texttt{[new-token]} is a very important part of the music.
& $\bullet$ 1. The \texttt{[new-token]} of the house was a very old man.
& \ac{metaicl-w} wins. CoLLEGe fails to use the word and generates a default sentence.
& \mbox{cello} + \mbox{bagpipe}
\\
\midrule
$\bullet$ 4. Allow the stock to settle then skim off the fat using \texttt{[new-token]} or fat separator.
$\bullet$ Add the pasta, cook it for about 12 minutes until just soft, then drain it in a \texttt{[new-token]}.
$\bullet$ Imagine you've got a bucketful of praise in every room of the house with a big \texttt{[new-token]}.
$\bullet$ Plants can be grown in tyres, old metal pots such as buckets, watering cans or even a \texttt{[new-token]}!
& $\bullet$ I'm going to put a \texttt{[new-token]} on the top of the stairs.
& $\bullet$ 1. The \texttt{[new-token]} is a small, round, flat, and smooth stone.
& Tie. The word refers to a kitchen utensil, likely a container, but neither model illustrates this. GPT-4o decides the \ac{metaicl-w} as the winner.
& \mbox{ladle} + \mbox{colander}
\\
\midrule
$\bullet$ At about the same age I was in a cafe and I spilt some pop on my \texttt{[new-token]}s.
$\bullet$ It's also the first International home \texttt{[new-token]} in history that is reversible as well.
$\bullet$ Such as how if you have sufficiently-stretchy pants you could take them off without first removing your \texttt{[new-token]}s.
$\bullet$ It was in a cinema on 86th Street where he sat in the balcony, wearing an old sports jacket and open-necked \texttt{[new-token]}.
& $\bullet$ I'm not sure if it's a good idea to wear \texttt{[new-token]} in the shower.
& $\bullet$ 1. The \texttt{[new-token]} is a very comfortable shoe.
& CoLLEGe wins. Both examples use the word as something wearable, but the thing is clearly not wearable in the shower.
& \mbox{trouser} + \mbox{shirt} \\
\bottomrule
\end{tabular}
\end{center}
\caption{New examples generated on study examples from the Chimera dataset \citep{Lazaridou2017MultimodalWM} by the \ac{metaicl-w} model and the CoLLEGe baseline on \mbox{Llama-2 7B}, with greedy decoding.
Like in Table~\ref{tab:babylm-generation}, the first author provides the judgment and reason (Judgment and Reason).
}
\label{tab:chimera-generation-llama-2}
\end{table*}

\begin{table*}[t]
\small
\begin{center}
\begin{tabular}{p{4cm}p{2.5cm}p{2.5cm}p{0.8cm}p{4cm}}
\toprule
\textbf{Study Example Sentences} & \textbf{\ac{metaicl-w}} & \textbf{Baseline} & \textbf{Word} & \textbf{Analysis} \\
\midrule
$\bullet$ the first blacksmiths were \texttt{[new-token]}.
$\bullet$ many civilisations were in the area that is now turkey, like the \texttt{[new-token]}, the roman empire and the byzantine empire.
$\bullet$ spread of hepatoscopy and astrology to \texttt{[new-token]}, etruscans, greeks and romans and to china
$\bullet$ the first major empire in the area was the \texttt{[new-token]} (from the 18th century to the 13th century bce).
&
1. the \texttt{[new-token]} were a people who lived in the area of turkey.
2. perhaps the most famous and widely used alchemical symbol, first popularized by \texttt{[new-token]} alchemists, is the ouroboros.
&
1. the \texttt{[new-token]} was a major power in the region, and its influence spread to the surrounding areas.
2. epistemology will suffer the \texttt{[new-token]} syndrome.
& hittites
&
The \ac{metaicl-w} model learns that \emph{hittites} is an ancient ethnic group. However, the greedy-decoded example copies the information (turkey) from the study example, while the sampled example makes seemingly plausible but factually incorrect generalizations (the earliest known ouroboros is found in ancient Egyptian text.)
The baseline also generates factual errors (Hittite plague exists but no ``hittites syndrome'').
\\
\midrule
$\bullet$ but, i know from his childhood, he says that his father always buys and \texttt{[new-token]} cars just as entertainment, i guess.
$\bullet$ in the end, yuji accepts shana's feelings and the two kiss, which activates a spell left to yuji by crimson lord lamia that \texttt{[new-token]} his existence, so yuji is no longer a torch.
$\bullet$ -- a medicine which \texttt{[new-token]} health and energy, as extract of malt, cod liver oil, etc.
$\bullet$ that diversifies energy supply, \texttt{[new-token]} ecosystems, protects biodiversity, and sequesters carbon.
&
1. the \texttt{[new-token]} of the \texttt{[new-token]} of the \texttt{[new-token]} of \textbf{[Infinite Repetition]}
2. he is a former prime minister, who is then \texttt{[new-token]} of his duties.
&
1. the \texttt{[new-token]} is a 3.5 mm jack, which is the standard for most headphones.
2. he \texttt{[new-token]} up to an incredible 2,620 sales per store in 2009, compared with the 60 to 80 that had been typical just a few years earlier.
& restores
&
For the \ac{metaicl-w} model, the greedy-decoded example is nonsense repetition (a common failure of LMs), while the sampled example uses the verb with incorrect syntax (``be restored to'') but pertinent semantics.
For the baseline, the greedy-decoded example also has incorrect syntax (use the word as a noun), and the use in the sampled example is semantically incorrect (the sales must have reached a similar level in the past).
\\
\midrule
$\bullet$ when i was a student \texttt{[new-token]}, i went to the prado in Madrid.
$\bullet$ \texttt{[new-token]} is hiking in the outdoors with a backpack full of all your food, clothing, water, and shelter.
$\bullet$ \texttt{[new-token]} may be done on short or long trips.
$\bullet$ \texttt{[new-token]} (wilderness)
&
1. \texttt{[new-token]} is a form of wilderness travel.
2. a small \texttt{[new-token]} campsite is usually composed of a place to sleep, food, tools and water.
&
1. \texttt{[new-token]} is a type of backpacking.
2. \texttt{[new-token]} is something all people can do.
& back-packing
&
The \ac{metaicl-w} model seems to infer the meaning of the word, but is rewording the study examples.
The baseline also seems to infer the meaning, but its sampled example is not very informative.
\\
\bottomrule
\end{tabular}
\end{center}
\caption{Error analysis of new examples generated from the BabyLM-10M test portion by the \ac{metaicl-w} model and the baseline. In each column of generated examples, the first one is generated by greedy decoding, and the second one by sampling with top-p=$0.92$.
}
\label{tab:babylm-generation-error}
\end{table*}

\begin{table*}[t]
\small
\begin{center}
\begin{tabular}{p{4cm}p{2.5cm}p{2.5cm}p{0.8cm}p{4cm}}
\toprule
\textbf{Study Example Sentences} & \textbf{\ac{metaicl-w}} & \textbf{CoLLEGe} & \textbf{Word} & \textbf{Analysis} \\
\midrule
$\bullet$ the first blacksmiths were \texttt{[new-token]}.
$\bullet$ many civilisations were in the area that is now turkey, like the \texttt{[new-token]}, the roman empire and the byzantine empire.
$\bullet$ spread of hepatoscopy and astrology to \texttt{[new-token]}, etruscans, greeks and romans and to china
$\bullet$ the first major empire in the area was the \texttt{[new-token]} (from the 18th century to the 13th century bce).
&
1. the \texttt{[new-token]} were a civilization in the area that is now turkey.
2. the \texttt{[new-token]} empire was destroyed by the medes and persians.
&
1. The \texttt{[new-token]} were a people of the ancient world.
2. 
& hittites
&
The \ac{metaicl-w} model learns that \emph{hittites} is an ancient ethnic group. However, the greedy-decoded example copies the information (turkey) from the study example, while the sampled example makes seemingly plausible but factually incorrect generalizations.
The CoLLEGe baseline's generation contains even less information, and the sampled result is empty (other sampled results not shown here are in Greek).
\\
\midrule
$\bullet$ but, i know from his childhood, he says that his father always buys and \texttt{[new-token]} cars just as entertainment, i guess.
$\bullet$ in the end, yuji accepts shana's feelings and the two kiss, which activates a spell left to yuji by crimson lord lamia that \texttt{[new-token]} his existence, so yuji is no longer a torch.
$\bullet$ -- a medicine which \texttt{[new-token]} health and energy, as extract of malt, cod liver oil, etc.
$\bullet$ that diversifies energy supply, \texttt{[new-token]} ecosystems, protects biodiversity, and sequesters carbon.
&
1. the \texttt{[new-token]} of the bacteria is a result of the action of the enzyme.
2. renewable energy \texttt{[new-token]} supply for energy.
&
1. The word '\texttt{[new-token]}' in example sentences.
2. Sample sentences from outside English.
& restores
&
For the \ac{metaicl-w} model, the greedy-decoded example uses the verb with incorrect part-of-speech but pertinent semantics, but the sampled example is correct and reasonable.
The CoLLEGe baseline fails to follow the instructions.
\\
\midrule
$\bullet$ when i was a student \texttt{[new-token]}, i went to the prado in Madrid.
$\bullet$ \texttt{[new-token]} is hiking in the outdoors with a backpack full of all your food, clothing, water, and shelter.
$\bullet$ \texttt{[new-token]} may be done on short or long trips.
$\bullet$ \texttt{[new-token]} (wilderness)
&
1. \texttt{[new-token]} is a type of outdoor recreation.
2. the song was first performed in 1933 by the hilda \texttt{[new-token]} choral society.
&
1. The \texttt{[new-token]} of the house was a very old man.
2. 10 pounds of \verb|___________| in ten days.
& back-packing
&
The \ac{metaicl-w} model correctly uses the word in the greedy-decoded example, but fails in the sampled example.
The baseline fails to understand the word and generates a default sentence in the greedy-decoded example.
\\
\bottomrule
\end{tabular}
\end{center}
\caption{Error analysis of new examples generated from the BabyLM-10M test portion by the \ac{metaicl-w} model and the CoLLEGe baseline with \mbox{Llama-2 7B}. In each column of generated examples, the first one is generated by greedy decoding, and the second one by sampling with top-p=$0.92$.
}
\label{tab:babylm-generation-error-llama-2}
\end{table*}

\clearpage
\section{Evaluation of Generated Definitions}
\label{app:definition-evaluation}

As we mentioned in Section~\ref{sec:definition-generation}, we use two definition generation datasets: CoLLEGe-DefGen \citep{Teehan2024CoLLEGeCE} and the Oxford test set \citep{gadetsky-etal-2018-conditional}.
The original datasets contain 954 and 12,232 words, from which we removed 4 and 2 duplicated words, respectively.
For CoLLEGe-DefGen, we keep the inflectional suffixes, such as ``-s'', ``-ed'', and ``-ly'', after the placeholder so that the placeholder only corresponds to the word stem. This is to remove the influence of morphological inflections.
Note that we use our placeholders instead of the \texttt{<nonce>} in the original text of CoLLEGe-DefGen.
In addition, we fixed several incorrect word/phrase replacements in the original dataset (for example, the phrase ``\emph{capital gains tax}'').
For the Oxford dataset, for simplicity and consistency with previous work, we do not keep the inflectional suffixes but rather replace the whole word with the placeholder.
There are 12\% examples in the Oxford test set in which we find no occurrences of any form of the word to be learned, but we keep them for consistency with previous work.

Additionally, as we also mentioned in Section~\ref{sec:definition-generation}, we have additional references of what can be achieved by specialized definition-generation models: the series of \mbox{FLAN-T5} \citep{chung2024scaling} models finetuned by \citet{giulianelli-etal-2023-interpretable} specifically on generating definitions.
This also follows what \citet{Teehan2024CoLLEGeCE} did.
These models were finetuned on three corpora, including the Oxford training set \citep{gadetsky-etal-2018-conditional}.
The series of finetuned \mbox{FLAN-T5} are listed on their GitHub page (\url{https://github.com/ltgoslo/definition_modeling?tab=readme-ov-file#definition-generation-models-for-english}) and can be accessed through Hugging Face model hub.
When evaluating the \mbox{FLAN-T5} models, a pseudo-word `\textit{wug}' is used as the placeholder for the new word, like in other off-the-shelf baselines (Section~\ref{sec:baseline-llama}) for a fair comparison. Each \mbox{FLAN-T5} model is prompted with an example sentence followed by a question, ``What is the definition of wug?'', as what \citet{giulianelli-etal-2023-interpretable} did.

Table~\ref{tab:definition-quantitative-1-shot-full} shows the full set of results of comparing the model-generated and ground-truth definitions from all models.
Table~\ref{tab:definition-quantitative-defgen} shows the average of 1-, 2-, and 3-shot results on the CoLLEGe-DefGen dataset.
Tables~\ref{tab:defgen-definition-more}~and~\ref{tab:oxford-definition} show additional definitions generated from the CoLLEGe-DefGen and Oxford test set by the baselines and the \ac{metaicl-w} models (in addition to Table~\ref{tab:defgen-definition} in Section~\ref{sec:definition-generation}).


\begin{table*}[t]
\small
\begin{center}
\begin{tabular}{p{2.7cm}l|cccc}
\toprule
\multicolumn{2}{c|}{\bf Model} & \multicolumn{2}{c}{\bf CoLLEGe-DefGen} & \multicolumn{2}{c}{\bf Oxford} \\
\bf Variant & \bf Method & \bf BERTScore F1 & \bf ROUGE-L & \bf BERTScore F1 & \bf ROUGE-L \\
\midrule
\multirow{2}{=}{Llama-3 8B}
           &             baseline          & 85.1 & 14.9 & 83.2 & 11.0 \\
           & +\ac{metaicl-w}               & 85.4 & 18.7 & \textbf{84.7} & \textbf{16.3} \\
\midrule
\multirow{2}{=}{Llama-3 8B Instruct}
                   &     baseline          & 85.3 & 17.6 & 83.6 & 12.5 \\
                   &+\ac{metaicl-w}        & \textbf{85.8} & \textbf{20.7} & \textbf{84.7} & \textbf{16.5} \\
\midrule\midrule
\multirow{3}{=}{Llama-2 7B}
                   &     baseline          & 84.4 & 14.7 & 83.9 & 13.0 \\
                   &+CoLLEGe               & 84.0 & 16.3 & 83.3 & 14.1 \\
                   &+\ac{metaicl-w}        & 82.9 & \textbf{18.0} & 83.6 & \textbf{15.6} \\
\midrule\midrule
FLAN-T5 Base  &+DefInstr baseline          & 83.1 & 13.1 & 84.4 & 16.5 \\
FLAN-T5 Large &+DefInstr baseline          & \textbf{83.8} & \textbf{15.5} & 84.7 & 17.4 \\
FLAN-T5 XL    &+DefInstr baseline          & 83.1 & 12.4 & \textbf{84.9} & \textbf{19.4} \\
\bottomrule
\end{tabular}
\end{center}
\caption{Quantitative evaluation of generated definitions by comparing them with ground-truth definitions.
This table extends Table~\ref{tab:definition-quantitative-1-shot} in the main text by adding additional results of the \mbox{Llama-2 7B} baseline and FLAN-T5 models.
No significant differences are found among \mbox{Llama-2 7B} models on BERTScore F1.
The FLAN-T5 models generally perform better than all other models on the Oxford dataset, but note that the Oxford dataset is in-distribution for these models, and these models may be overfitting to this dataset (see Table~\ref{tab:oxford-definition} for examples and discussion).
}
\label{tab:definition-quantitative-1-shot-full}
\end{table*}

\begin{table*}[t]
\small
\begin{center}
\begin{tabular}{p{2.7cm}l|cc}
\toprule
\multicolumn{2}{c|}{\bf Model} & \multicolumn{2}{c}{\bf CoLLEGe-DefGen} \\
\bf Variant & \bf Method & \bf BERTScore F1 & \bf ROUGE-L \\
\midrule
\multirow{2}{=}{Llama-3 8B}
& baseline                 & 85.8 & 17.8 \\
& +\ac{metaicl-w}          & 85.9 & 21.1 \\
\midrule
\multirow{2}{=}{Llama-3 8B Instruct}
& baseline                 & 85.9 & 19.5 \\
&+\ac{metaicl-w}           & \textbf{86.2} & \textbf{22.6} \\
\midrule
\multirow{3}{=}{Llama-2 7B}
& baseline                 & 85.2 & 17.0 \\
&+\ac{metaicl-w}           & 84.0 & 19.9 \\
&+CoLLEGe                  & 84.2 & 16.9 \\
\bottomrule
\end{tabular}
\end{center}
\caption{Quantitative evaluation of generated definitions by comparing them with ground-truth definitions in the CoLLEGe-DefGen dataset. Definitions are generated 1-, 2-, and 3-shot and scores are averaged. All definitions are generated with greedy decoding. For models finetuned with \ac{metaicl-w}, scores are averaged across 3 runs.
CoLLEGe* results are from Table~2 of \citet{Teehan2024CoLLEGeCE}, which is based on \mbox{Llama-2 7B} and slightly different data processing (see Appendix~\ref{app:definition-evaluation}). We do not have \mbox{FLAN-T5} models here since \citet{giulianelli-etal-2023-interpretable} finetuned them to use only one usage example.}
\label{tab:definition-quantitative-defgen}
\end{table*}

\begin{table*}[t]
\small
\begin{center}
\begin{tabular}{p{3.15cm}p{3cm}p{3.15cm}p{3.15cm}p{1.4cm}}
\toprule
\bf Example Sentence & \bf True Definition & \bf \ac{metaicl-w} & \bf Baseline & \bf Word \\
\midrule
\multirow{2}{=}{As the hurricane neared, the residents began to \texttt{[new-token]} their windows to protect their homes from the impending storm.}
& \multirow{2}{=}{to cover or seal windows, doors, or other openings of a building with boards, typically to protect it from damage or unauthorized entry.}
& to protect from harm or danger
& \textbf{to prepare for a hurricane by boarding up windows}
& board up
\\
\cmidrule{3-4}
&
& to make something more secure or safe by covering it with a layer of material
& \textbf{to secure or fasten something, especially a window, to prevent it from being damaged or destroyed}
\\
\midrule
\multirow{2}{=}{The gentle hum of the air conditioner provided a \texttt{[new-token]} soundtrack for her midday nap.}
& \multirow{2}{=}{having a calming or relieving effect, especially in terms of reducing pain or discomfort.}
& \textbf{a sound that is not loud enough to be heard}
& a small, furry, brown creature that lives in trees.
& soothing
\\
\cmidrule{3-4}
&
& \textbf{a soothing, calming, or quiet sound}
& a wug is a word that is not yet known to the speaker, but is assumed to be a real word.
\\
\midrule
\multirow{2}{=}{In their groundbreaking research, the team of geneticists successfully deactivated the \texttt{[new-token]}, resulting in the unexpected bloom of dormant traits within the lab mice.}
& \multirow{2}{=}{a type of gene that codes for a protein, known as a repressor, which inhibits the expression of one or more genes by binding to the operator or associated silencers.}
\vspace{0.5cm}
& \textbf{a gene that is turned off in a cell}
& a hypothetical new word that does not yet exist in the English language.
& repressor gene
\\
\cmidrule{3-4}
&
& \textbf{a gene or set of genes that controls the development of a particular trait or characteristic}
& a hypothetical word used in linguistic research to test the ability to form and use new words.
\\
\midrule
\multirow{2}{=}{She preferred the \texttt{[new-token]} wilderness to the stifling orderliness of city life.}
& \multirow{2}{=}{not restricted or limited; free; unconfined.}
& not having a definite shape or form
& a small, furry animal
& untrammeled
\\
\cmidrule{3-4}
&
& a place where there are many trees, especially in a forest or a park
& \textbf{a mythical creature that is half-wolf and half-bear}
\\
\midrule
\multirow{2}{=}{In the heart of her rustic kitchen, Grandma carefully seasoned the \texttt{[new-token]}, her secret ingredient for the family's cherished Sunday stew.}
& \multirow{2}{=}{The chest portion of a young cow, typically used in cooking for its tender meat.}
& \textbf{a mixture of herbs and spices used to flavor food}
& a mythical creature that resembles a cross between a dog and a frog.
& breast of veal
\\
\cmidrule{3-4}
&
& \textbf{a small, usually round, piece of food, especially a piece of meat or a vegetable, cut off from a larger piece and cooked separately}
& a type of meat
\\
\midrule
\multirow{2}{=}{The contractor recommended \texttt{[new-token]} for insulation due to its excellent thermal resistance and fireproofing properties.}
& \multirow{2}{=}{a type of insulation material made from melted rock or industrial waste that is spun into a fibrous structure. It is used for thermal insulation, fire protection, and soundproofing in both residential and commercial buildings.}
\vspace{1.95cm}
& \textbf{a material used to insulate a building}
& a unit of insulation used in construction
& mineral wool
\\
\cmidrule{3-4}
&
& \textbf{a material used to prevent heat transfer, especially in buildings}
& a type of insulation material
\\
\bottomrule
\end{tabular}
\end{center}
\caption{Definitions generated by both the pre-trained and instruction-tuned variant of \mbox{Llama-3 8B} (baselines) and the models finetuned from them with \ac{metaicl-w} on BabyLM-10M with greedy decoding, using the prompt ``The word \texttt{[new-token]} in the above sentence(s) is defined as "'' (so we can extract continuations before the closing quote as the generated definitions).
Each definition is generated using the single example sentence shown and provided in context.
The pre-trained/instruction-tuned variant is shown at the top/bottom of each row of example.
We boldface the winner judged by \mbox{GPT-4o}.
You may observe the quality of instruction-tuned variant is better than the pre-trained variant by manually comparing their definition in each example.}
\label{tab:defgen-definition-more}
\end{table*}

\begin{table*}[t]
\small
\begin{center}
\begin{tabular}{p{2.1cm}p{2cm}p{2cm}p{2cm}p{2cm}p{2cm}p{1cm}}
\toprule
\bf Example Sentence & \bf True Definition & \bf \ac{metaicl-w} & \bf Baseline & \bf CoLLEGe & \bf FLAN-XL-DefInstr & \bf Word \\
\midrule
many people in the world have to \texttt{[new-token]} on \$ 1 a day
& support oneself
& to live on something, especially food, in a way that is not very good or healthy
& to struggle to make ends meet
& the \texttt{[new-token]} of two numbers.
& live on
& subsist
\\
\midrule
this food \texttt{[new-token]} the patient
& restore strength
& to make something more intense or strong
& a type of food
& to make sick or ill.
& to give strength
& revitalizes
\\
\midrule
he is always well-\texttt{[new-token]}
& care for one's external appearance
& well - dressed
& a type of wug
& to make a person feel more comfortable or at ease.
& in good health
& groomed
\\
\midrule
the lawyers tried to \texttt{[new-token]} the credibility of the witnesses
& challenge the honesty or veracity of
& to make something more convincing or believable
& to question the credibility of a witness
& to \texttt{[new-token]} (someone) with a blow or \texttt{[new-token]} (something) by a blow.
& to challenge the honesty or veracity of
& impeach
\\
\midrule
the car squeaks to a halt and she glares at him because of his \texttt{[new-token]} stop.
& characterized by abrupt stops and starts
& a sudden, sharp, high - pitched sound, especially one made by a car's brakes or a bird's call
& a made-up word
& a sudden, \texttt{[new-token]}, or \texttt{[new-token]} attack of pain.
& a jerk that causes an object to move abruptly
& jerky
\\
\midrule
try the full plate pork \texttt{[new-token]} : tender pork, oregano-spiked greek salad, warm puffy pita, rice, and aromatic tzatziki-topped lemon potatoes.
& a greek dish of pieces of meat grilled on a skewer
& a dish of meat, usually pork, served with a sweet and sour sauce, and often served with rice and vegetables
& a type of dish that is a combination of pork, rice, and potatoes, typically served with a side of salad and pita bread.
& a dish of meat, fish, or vegetables cooked in a sauce.
& a greek dish of grilled meat served in a pita .
& souvlaki
\\
\midrule
extend the tv antenna \textbf{(word is absent)}
& extend or stretch out to a greater or the full length
& a small, usually round, piece of metal or plastic used to connect two wires together
& a type of bird
& to \texttt{[new-token]} or \texttt{[new-token]} (a person) with a weapon.
& raise or extend vertically
& stretch
\\
\midrule
the red light gave the central figure increased emphasis \textbf{(word is absent)}
& special importance or significance
& a red light
& a wug is a wug
& a sudden, violent, and often uncontrollable attack of fear, dread, or apprehension.
& special importance or significance
& accent
\\
\bottomrule
\end{tabular}
\end{center}
\caption{Definitions generated by the instruction-tuned variant of \mbox{Llama-3 8B} (baseline), the \ac{metaicl-w} model finetuned from it with greedy decoding, the CoLLEGe model, and \mbox{FLAN-XL-DefInstr} (i.e., \mbox{FLAN-T5} XL +DefInstr baseline), using the prompt ``The word \texttt{[new-token]} in the above sentence(s) is defined as "'' (\texttt{[new-token]} can be replaced by other placeholders, as we mentioned in Section~\ref{sec:definition-generation}).
Each definition is generated using the single example sentence shown and provided in context.
The \ac{metaicl-w} model generates reasonable definitions given the context, but is often much longer than the ground-truth definitions, likely because it is not fitted to this dataset.
The \mbox{Llama-3 8B} baseline is often generating low-quality or repetitive definitions, and sometimes sticks to its prior knowledge of the pseudo-word ``\emph{wug}.''
CoLLEGe often generates definitions that contain the \texttt{[new-token]}, or fail to understand the word correctly.
FLAN-XL-DefInstr generates definitions pretty close to the ground-truth, but is sometimes suspicious of overfitting to or memorizing the data, as its definition for `impeach' and `accent' (absent in the example) may suggest.}
\label{tab:oxford-definition}
\end{table*}

\clearpage\clearpage
\section{Concepts of ``Word''}
\label{app:word}
The term ``word'' can refer to linguistic units with nuanced variations.
Here, we describe the concepts of ``word'' in different contexts of the paper and their implications.
Surprisingly, our models are somehow robust to these variations of ``word.'' Future work may further improve the processing of words and conduct targeted evaluations of morphological variations of the learned words.

\paragraph{Word usage datasets}
In the two datasets we constructed for training and finetuning (Section~\ref{sec:dataset} and Appendix~\ref{app:dataset}), a ``word'' means a word-form, which is instantiated as an individual token extracted from the word-level tokenization (using spaces and punctuations as boundaries).
Therefore, for the same lexeme, a sentence using one of its word-form is not considered an example of another word-form. For instance, a sentence using other inflected forms of ``\emph{ski}'' like ``\emph{Susie likes skiing fast down the snowy mountain on her new skis}'' is not included in the example set of ``\emph{ski}.''
Meanwhile, when two word-forms of the same lexeme occur in one sentence, meta-learning one of the word-form could be easier since the other word-form may not be masked. For instance, ``\emph{skis}'' in the sentence ``\emph{I saw Susie ski fast down the snowy mountain on her new skis}'' could make it easier to guess the word ``\emph{ski}.''
In our work, we focus on learning word-forms, but if we aim to learn a lexeme, this case will reveal the identity of the lexeme we try to mask, undermining our effort on the novelty of the learned word.
On the other hand, a word-form in different syntactic categories is considered the same word, and the usage examples will be mixed together regardless of the syntactic categories. Such words are rare, but they introduce syntactic uncertainties in word learning. Syntactic uncertainties are natural, but may increase the difficulty of learning.

\paragraph{Pseudo-words}
In our off-the-shelf baselines (Section~\ref{sec:baseline-llama} and the additional specialized \mbox{FLAN-T5} models in Section~\ref{sec:definition-generation}) and comparison of generations (Appendix~\ref{app:comparing-generations}), we replace the word to learn by a pseudo-word, like ``\emph{dax}'' or ``\emph{wug}'', regardless of the word's syntactic category and other aspects of meaning.
The pseudo-word is then tokenized, usually by a subword tokenizer for LLMs (thus may have multiple tokens).
We choose the pseudo-word to be meaningless and commonly used in linguistic tests.
However, a pre-trained LLM like \mbox{Llama} may have priors of certain aspects of the pseudo-word's meaning based on its form.
One aspect of the meaning is syntax.
For example, from the sentence ``\emph{Susie goes skiing in the winter}'', we replace ``\emph{skiing}'' with ``\emph{dax}'' and have the sentence ``\emph{Susie goes dax in the winter}.'' The sentence has a problem: the part of speech of ``\emph{skiing}'' is gerund, but ``\emph{dax}'' does not look like a gerund (since it does not end in ``\emph{-ing}''). So the sentence could mislead an LLM like \mbox{Llama}, which can use morphological information from its subword tokenization.
Another aspect of the meaning is semantics.
For example, in Table~\ref{tab:oxford-definition}, the baseline model sometimes sticks to its prior knowledge of the pseudo-word ``\emph{wug},'' as reflected in its generated definitions like ``\emph{a made-up word}'' and ``\emph{a type of bird}'' (``\emph{wug}'' referred to a bird-like creature in the Wug Test of \citealp{Berko1958TheCL}).
We admit that this problem may weaken our baselines and comparison of generations.
Future work should use more suitable pseudo-words, preserving the morphological inflections while removing the semantic information.

\paragraph{Evaluation datasets}
Words to be learned in the Chimera, CoLLEGe-DefGen, and Oxford datasets are lexemes, so examples of each word use (different) inflected word-forms.
To ensure the placeholder consistently represents the same text, we replace only the word stem with the placeholder and retain the inflectional suffixes in the original word-forms on the Chimera and CoLLEGe-DefGen datasets. (We still replace word-forms in Oxford to make our practice consistent with previous ones.)
In addition, words to be learned in the CoLLEGe-DefGen dataset also include multiwords or phrases, like the ``\emph{categorical imperative}'' example in Table~\ref{tab:defgen-definition}.
See Appendix~\ref{app:definition-evaluation} for further details of preprocessing.
Surprisingly, although our placeholder token represents a word-form in the \mbox{BabyLM-10M} dataset we constructed, \ac{metaicl-w} models finetuned on \mbox{BabyLM-10M} still perform well when using the token to represent a word stem in these datasets.

\clearpage\clearpage
\section{Changes in Other Capabilities}
\label{app:other_capabilities}
How does \ac{metaicl-w} finetuning affect other capabilities of language models?
As we mentioned in Section~\ref{sec:finetuning}, we finetune only the input and output embeddings of the two special tokens while freezing all other model parameters. Therefore, we expect that \ac{metaicl-w} finetuning will not change other general capabilities of the finetuned language model.
To validate this, we evaluate the pre-trained \mbox{Llama-3 8B} and the \ac{metaicl-w} finetuned from it on the BLiMP benchmark \citep{warstadt-etal-2020-blimp-benchmark}, which evaluates the grammatical capabilities of language models.
Results are shown in Table~\ref{tab:blimp}.
We find that \ac{metaicl-w} finetuning does not change the accuracies very much on most subsets in the benchmark: Most accuracies does not change or change within 0.3\%, except for the subset matrix\_question\_npi\_licensor\_present, which has a 9.7\% decrease and high variance in accuracy.
These results reflect that other capabilities of language models are almost unaffected by \ac{metaicl-w}.

\begin{table*}[t]
\small
\begin{center}

\begin{tabular}{llrr}
\toprule
Phenomenon & UID & Llama-3 8B & +\ac{metaicl-w} \\
\midrule
\multirow[t]{2}{*}{anaphor agreement} & anaphor\_gender\_agreement & 98.9 & 98.9(0.0) \\
 & anaphor\_number\_agreement & 99.5 & 99.5(0.0) \\
\cline{1-4}
\multirow[t]{9}{*}{argument structure} & animate\_subject\_passive & 80.7 & 80.7(0.1) \\
 & animate\_subject\_trans & 84.2 & 84.1(0.2) \\
 & causative & 76.7 & 76.7(0.0) \\
 & drop\_argument & 79.8 & 79.8(0.0) \\
 & inchoative & 70.6 & 70.6(0.0) \\
 & intransitive & 83.7 & 83.7(0.0) \\
 & passive\_1 & 90.5 & 90.5(0.0) \\
 & passive\_2 & 90.8 & 90.8(0.0) \\
 & transitive & 90.1 & 90.1(0.0) \\
\cline{1-4}
\multirow[t]{7}{*}{binding} & principle\_A\_c\_command & 80.3 & 80.3(0.0) \\
 & principle\_A\_case\_1 & 100.0 & 100.0(0.0) \\
 & principle\_A\_case\_2 & 93.8 & 93.9(0.1) \\
 & principle\_A\_domain\_1 & 99.3 & 99.3(0.0) \\
 & principle\_A\_domain\_2 & 88.3 & 88.3(0.0) \\
 & principle\_A\_domain\_3 & 52.8 & 52.7(0.1) \\
 & principle\_A\_reconstruction & 45.3 & 45.3(0.0) \\
\cline{1-4}
\multirow[t]{5}{*}{control/raising} & existential\_there\_object\_raising & 85.0 & 85.0(0.0) \\
 & existential\_there\_subject\_raising & 89.9 & 89.9(0.0) \\
 & expletive\_it\_object\_raising & 80.3 & 80.3(0.0) \\
 & tough\_vs\_raising\_1 & 68.8 & 68.8(0.0) \\
 & tough\_vs\_raising\_2 & 87.2 & 87.2(0.0) \\
\cline{1-4}
\multirow[t]{8}{*}{determiner-noun agreement} & determiner\_noun\_agreement\_1 & 99.5 & 99.5(0.0) \\
 & determiner\_noun\_agreement\_2 & 99.0 & 99.0(0.0) \\
 & determiner\_noun\_agreement\_irregular\_1 & 96.9 & 96.9(0.0) \\
 & determiner\_noun\_agreement\_irregular\_2 & 96.8 & 96.8(0.0) \\
 & determiner\_noun\_agreement\_with\_adj\_1 & 97.5 & 97.5(0.0) \\
 & determiner\_noun\_agreement\_with\_adj\_2 & 95.4 & 95.4(0.0) \\
 & determiner\_noun\_agreement\_with\_adj\_irregular\_1 & 92.5 & 92.5(0.0) \\
 & determiner\_noun\_agreement\_with\_adj\_irregular\_2 & 94.9 & 94.9(0.0) \\
\cline{1-4}
\multirow[t]{2}{*}{ellipsis} & ellipsis\_n\_bar\_1 & 79.6 & 79.6(0.0) \\
 & ellipsis\_n\_bar\_2 & 92.7 & 92.7(0.1) \\
\cline{1-4}
\multirow[t]{7}{*}{filler gap} & wh\_questions\_object\_gap & 81.9 & 81.9(0.0) \\
 & wh\_questions\_subject\_gap & 91.7 & 91.7(0.0) \\
 & wh\_questions\_subject\_gap\_long\_distance & 88.0 & 88.1(0.0) \\
 & wh\_vs\_that\_no\_gap & 97.2 & 97.2(0.0) \\
 & wh\_vs\_that\_no\_gap\_long\_distance & 95.6 & 95.6(0.0) \\
 & wh\_vs\_that\_with\_gap & 39.9 & 39.8(0.1) \\
 & wh\_vs\_that\_with\_gap\_long\_distance & 31.7 & 31.7(0.0) \\
\cline{1-4}
\multirow[t]{2}{*}{irregular forms} & irregular\_past\_participle\_adjectives & 95.5 & 95.5(0.0) \\
 & irregular\_past\_participle\_verbs & 88.2 & 88.2(0.0) \\
\cline{1-4}
\multirow[t]{8}{*}{island effects} & adjunct\_island & 88.5 & 88.5(0.0) \\
 & complex\_NP\_island & 63.2 & 63.1(0.1) \\
 & coordinate\_structure\_constraint\_complex\_left\_branch & 72.0 & 72.0(0.1) \\
 & coordinate\_structure\_constraint\_object\_extraction & 85.7 & 85.7(0.0) \\
 & left\_branch\_island\_echo\_question & 42.6 & 42.5(0.2) \\
 & left\_branch\_island\_simple\_question & 84.8 & 84.8(0.1) \\
 & sentential\_subject\_island & 50.2 & 50.3(0.0) \\
 & wh\_island & 79.7 & 79.7(0.0) \\
\cline{1-4}
\multirow[t]{7}{*}{npi licensing} & matrix\_question\_npi\_licensor\_present & 81.7 & 72.0(8.7) \\
 & npi\_present\_1 & 61.6 & 61.6(0.0) \\
 & npi\_present\_2 & 70.1 & 70.1(0.0) \\
 & only\_npi\_licensor\_present & 97.8 & 97.5(0.2) \\
 & only\_npi\_scope & 88.5 & 88.2(0.3) \\
 & sentential\_negation\_npi\_licensor\_present & 99.5 & 99.5(0.0) \\
 & sentential\_negation\_npi\_scope & 71.4 & 71.4(0.0) \\
\cline{1-4}
\multirow[t]{4}{*}{quantifiers} & existential\_there\_quantifiers\_1 & 98.7 & 98.7(0.0) \\
 & existential\_there\_quantifiers\_2 & 67.1 & 67.1(0.0) \\
 & superlative\_quantifiers\_1 & 94.2 & 94.2(0.0) \\
 & superlative\_quantifiers\_2 & 90.0 & 90.2(0.3) \\
\cline{1-4}
\multirow[t]{6}{*}{subject-verb agreement} & distractor\_agreement\_relational\_noun & 87.5 & 87.6(0.0) \\
 & distractor\_agreement\_relative\_clause & 73.7 & 73.7(0.1) \\
 & irregular\_plural\_subject\_verb\_agreement\_1 & 92.1 & 92.1(0.0) \\
 & irregular\_plural\_subject\_verb\_agreement\_2 & 94.4 & 94.4(0.0) \\
 & regular\_plural\_subject\_verb\_agreement\_1 & 94.3 & 94.3(0.0) \\
 & regular\_plural\_subject\_verb\_agreement\_2 & 93.6 & 93.6(0.0) \\
\cline{1-4}
NaN & Mean & 83.5 & 83.3(0.1) \\
\bottomrule
\end{tabular}

\caption{Accuracies on BLiMP \citep{warstadt-etal-2020-blimp-benchmark}.
We show the mean and the standard deviation (in brackets) of 3 runs of \ac{metaicl-w}.
\ac{metaicl-w} accuracies are very close to the pre-trained model.}
\label{tab:blimp}
\end{center}
\end{table*}


\end{document}